\PassOptionsToPackage{hyphens}{url}
\documentclass[11pt]{article}
\usepackage[preprint]{acl}
\usepackage{times}
\usepackage{latexsym}
\usepackage[T1]{fontenc}
\usepackage[utf8]{inputenc}
\usepackage{graphicx}
\usepackage{booktabs}
\usepackage{multirow}
\usepackage{array}
\usepackage{tabularx}
\usepackage{xcolor}

\usepackage{amsmath}
\usepackage{enumitem}
\usepackage{placeins}
\usepackage{float}
\usepackage{algorithm}
\usepackage{algorithmic}
\usepackage{listings}
\usepackage{tcolorbox}
\tcbuselibrary{breakable,skins}
\usepackage{etoolbox}
\makeatletter
\newcommand{\promptneedspace}[1]{%
  \par\begingroup
  \dimen@=#1\relax
  \advance\dimen@\pagetotal
  \ifdim\dimen@>\pagegoal\pagebreak\fi
  \endgroup
}
\makeatother
\newsavebox{\autotabularbox}
\let\origbackslashtabular\tabular
\let\endorigbackslashtabular\endtabular
\renewenvironment{tabular}[2][]{%
  \begin{lrbox}{\autotabularbox}%
  \ifstrempty{#1}{\origbackslashtabular{#2}}{\origbackslashtabular[#1]{#2}}%
}{%
  \endorigbackslashtabular
  \end{lrbox}%
  \ifdim\wd\autotabularbox>\linewidth
    \resizebox{\linewidth}{!}{\usebox{\autotabularbox}}%
  \else
    \usebox{\autotabularbox}%
  \fi
}
\makeatletter
\g@addto@macro\UrlBreaks{\do\/\do\-\do\_\do\.\do\:\do\?\do\&\do\=\do\#}
\makeatother
\AtBeginEnvironment{thebibliography}{\sloppy}
\lstdefinestyle{promptstyle}{
  basicstyle=\scriptsize,
  breaklines=true,
  breakatwhitespace=false,
  columns=fullflexible,
  keepspaces=true,
  showstringspaces=false,
  upquote=true,
  frame=none,
  aboveskip=0pt,
  belowskip=0pt,
}
\newtcolorbox{promptbox}[1]{
  enhanced,
  breakable,
  boxrule=0.5pt,
  arc=1mm,
  colback=black!1,
  colframe=black!35,
  colbacktitle=black!6,
  coltitle=black,
  fonttitle=\small\bfseries,
  title={#1},
  left=1.2mm,
  right=1.2mm,
  top=0.8mm,
  bottom=0.8mm,
}
\newtcolorbox{tracecase}[1]{
  enhanced,
  boxrule=0.55pt,
  sharp corners,
  colback=white,
  colframe=black!80,
  colbacktitle=black!5,
  title={#1},
  fonttitle=\bfseries\footnotesize,
  coltitle=black,
  left=1.2mm,
  right=1.2mm,
  top=0.8mm,
  bottom=0.8mm,
  before skip=2pt,
  after skip=2pt,
  before upper={\raggedright\sloppy},
}
\newenvironment{promptstrip}[1]{%
  \begin{promptbox}{#1}
}{%
  \end{promptbox}
}

\title{Lost in Implicit Routing:\ How LLM Agents Fail to Load Relevant Skills at Scale}
\author{
  Jinlin Liu\textsuperscript{1} \quad
  Wei Chen\textsuperscript{2}\thanks{Corresponding authors.} \quad
  Xiaojin Zhang\textsuperscript{2}\footnotemark[1] \\
  \textsuperscript{1}The University of Hong Kong \\
  \textsuperscript{2}Huazhong University of Science and Technology
}
\hypersetup{pdfauthor={Jinlin Liu, Wei Chen, Xiaojin Zhang}}

\begin{document}
\maketitle

\begin{abstract}
Reusable skills let LLM agents load task-specific instructions on demand, but they make execution depend on implicit routing over skill metadata.  We test whether this design scales by holding SkillsBench tasks and their golden skills fixed while adding irrelevant skills.  We find \emph{skill routing collapse}: agents increasingly load fewer or no skills, even though the relevant skills remain available.  With 1,000 irrelevant skills, loaded-skill recall falls from 73.6\% to 27.4\% and the zero-load rate rises from 9.3\% to 51.6\%, whereas explicit selection from the same library retains 73.9\% recall.  Loading other skills, including plausible alternatives to golden skills, is too rare to explain this decline.  Masking only the descriptions of irrelevant skills largely restores skill loading, yet with the metadata intact, explicitly asking agents to plan skill use still yields lower recall than preselection.  These findings point to semantic interference from descriptions and instruction interference when skill choice competes with task execution.  We also evaluate explicit preselection as a practical mitigation, including its coverage and latency trade-offs.  Our analysis informs skill protocols and evaluation for agents with growing skill libraries.\footnote{Code: \url{https://github.com/jinlinll/lost-in-implicit-routing}.}
\end{abstract}

\section{Introduction}
Large language model (LLM) agents now operate tools and files to complete multi-step workflows in software engineering, data analysis, document processing, and personal assistance \citep{yao2023react,liu2023agentbench,wang2024agentSurvey}.  Such agents need reusable procedures: a plotting routine, code-audit workflow, spreadsheet operation, or domain-specific calculation should not be rediscovered in every new task.

\begin{figure}[!t]
\centering
\includegraphics[width=\columnwidth]{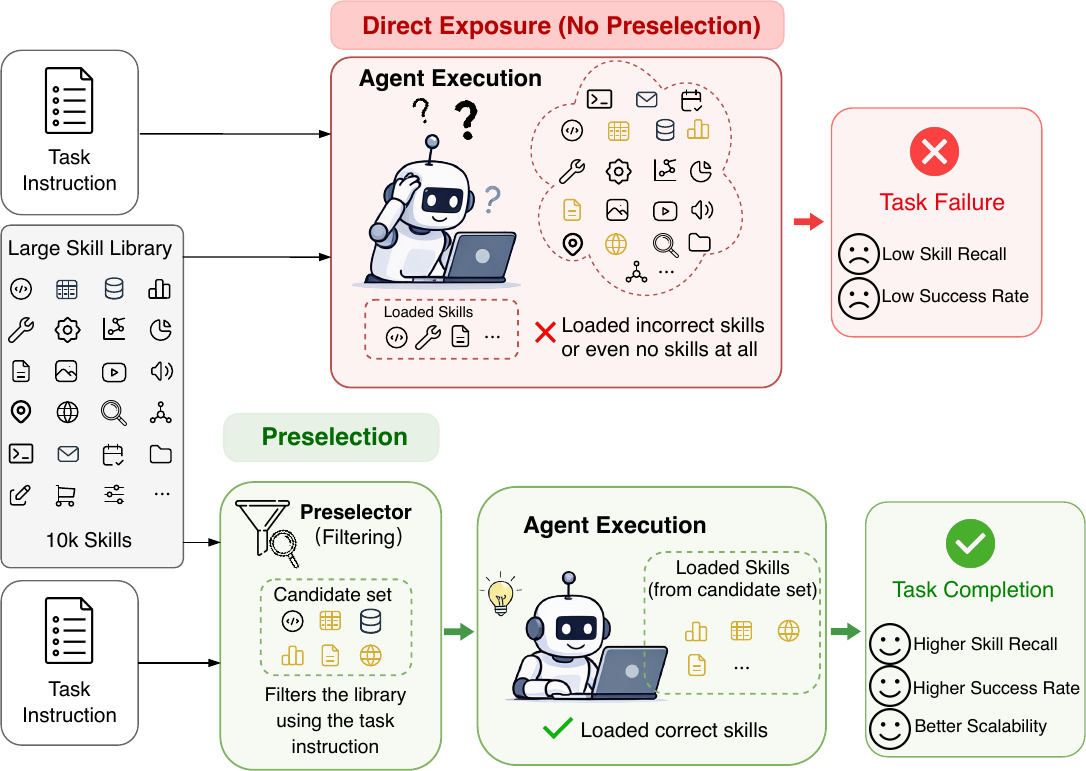}
\caption{Lost in implicit routing.  The relevant skills remain installed, but as irrelevant skills grow, agents increasingly fail to load them during implicit execution.  Explicit routing recovers much of the missing skill recall.}
\label{fig:teaser-skills-scaling}
\end{figure}

Agent skills package such procedures and their supporting resources into reusable modules \citep{anthropicAgentSkills2025,anthropicEquippingSkills2025}, and public skill libraries are growing quickly: one analysis reports 2,179 listed skills on January 16, 2026 and 40,285 on February 5 \citep{ling2026agentSkills}.  Skill selection is therefore becoming a scaling problem.

We use \emph{skills} to mean Anthropic's Agent Skills, whose progressive-disclosure design works as follows: the model first sees only skill names and descriptions, and opens a full skill file only if it judges the skill relevant \citep{anthropicSkills2026}.  Routing is thus not a separate stage: the agent decides whether and which skills to load while planning actions, inspecting files, and solving the task.  We call this execution-time selection \emph{implicit routing}.  It keeps long skill files out of context, but assumes that an agent can pick out the few relevant skills from many descriptions while simultaneously solving the task.

We test this assumption on SkillsBench \citep{li2026skillsbench}, which provides fixed task instructions, workspaces, verifiers, and relevant skills; for each task, we keep these fixed and vary only the number of added irrelevant skills.  The result is \emph{skill routing collapse}: the right skills are installed, but agents increasingly load fewer skills or none at all.  As irrelevant skills grow from 0 to 1,000, loaded-skill recall under implicit routing drops from 73.6\% to 27.4\% and the zero-load rate rises from 9.3\% to 51.6\%.  Agents occasionally load non-golden alternatives, but too rarely to explain the collapse.  Yet the same models retain 73.9\% selected-skill recall when asked only to choose skills from the full library.  Relevant skills thus remain recognizable; they are lost when routing becomes a subtask of execution.

End-to-end Pass@3 falls less sharply, from 51.6\% to 38.6\%, because strong agents can sometimes solve around missing skills; task success alone can therefore make a failing skill system look healthy.  We make three contributions:
\begin{itemize}[leftmargin=1.5em,itemsep=0.2em,topsep=0.2em]
    \item We identify and measure \emph{skill routing collapse}: growing libraries make agents load fewer or no skills, rather than substitute other skills for golden ones.
    \item We compare LLM, retrieval, and hybrid preselection on libraries of up to 10,000 added skills, and derive scale-dependent recommendations together with their token, coverage, and latency costs.
    \item We show that irrelevant descriptions suppress skill loading and that selecting skills while solving the task makes selection less reliable, revealing semantic and instruction interference beyond context growth alone.
\end{itemize}

\begin{figure*}[t]
\centering
\includegraphics[width=\textwidth]{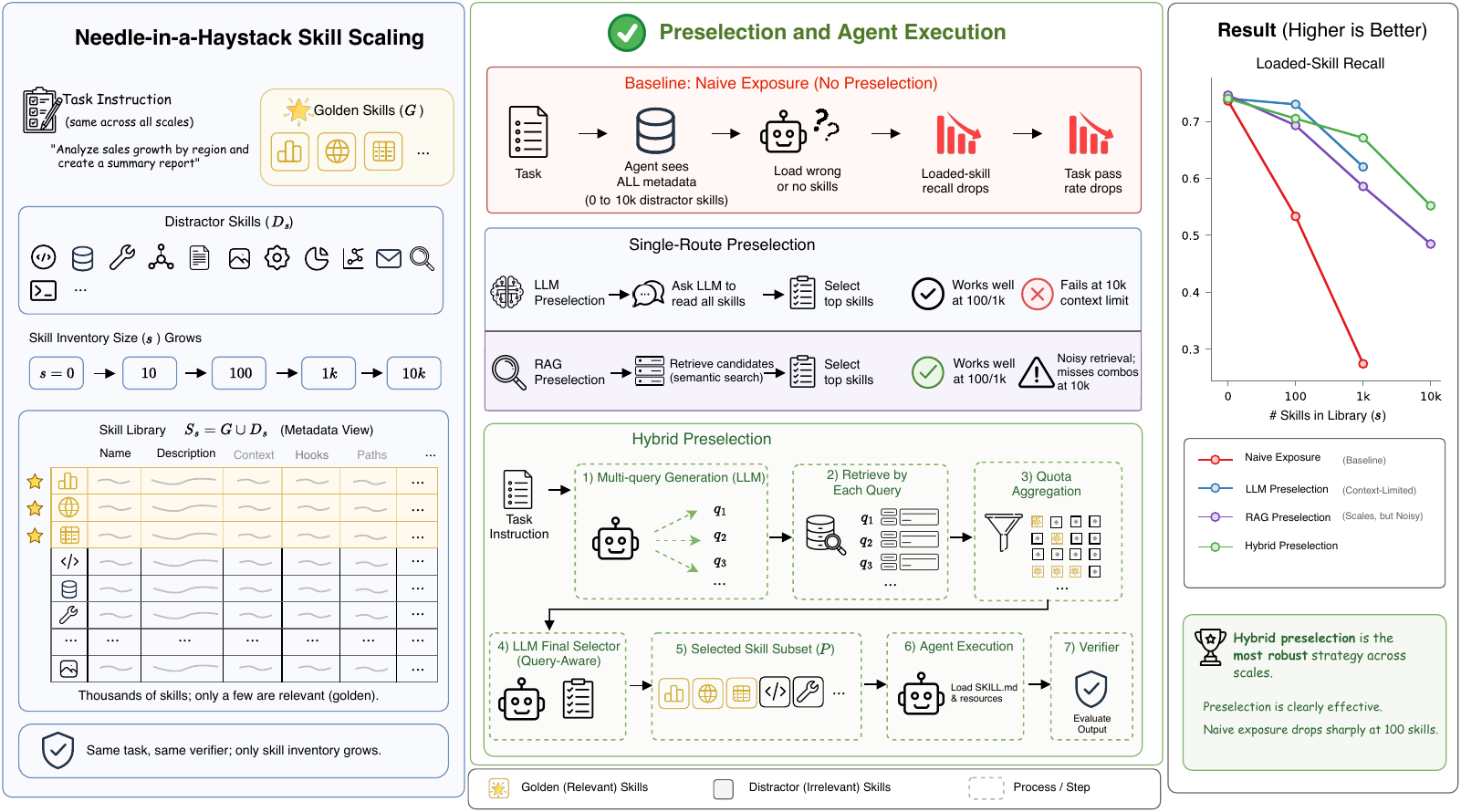}
\caption{Overview of the controlled skill-scaling setup and routing strategies.  Starting from an existing SkillsBench task, we keep the instruction, files, verifier, and relevant skills fixed while varying only the number of added irrelevant skills.}
\label{fig:overview-skills-scaling}
\end{figure*}

\section{A Controlled Test for Skill Routing}
\label{sec:controlled-test}
\subsection{Task setup}
We represent each task as $\tau=(I,F,V,G)$: instruction $I$, workspace $F$, executable verifier $V$, and the set $G$ of relevant (golden) SkillsBench skills.  A run succeeds only if $V$ passes after the agent finishes; no LLM judge is used.  With $s$ added irrelevant skills, the installed library is $S_s=G\cup D_s$, where $D_s$ is sampled independently of the task.  Varying only $s$ tests whether an agent can still find and load the needed skills as the metadata list grows (Figure~\ref{fig:overview-skills-scaling}).

\subsection{Routing strategies}
Under \emph{implicit routing}, the executor sees the metadata of all skills in $S_s$ and decides during execution whether to open any skill.  \emph{Explicit routing} adds a preselection step: a preselector receives the task and the metadata and exposes only a compact subset to the executor, which still follows progressive disclosure and must open a skill before using it.  We compare three preselectors.  \emph{LLM preselection} chooses directly from the full metadata list; \emph{RAG preselection} retrieves skills by embedding similarity; and \emph{hybrid preselection} retrieves candidates and lets an LLM make the final choice (pseudocode in Appendix~\ref{app:algorithms}).  We also include a no-skill baseline.

\paragraph{Diagnosis versus mitigation.}  We use explicit routing in two ways.  In \emph{routing-only diagnosis}, the selector sees the task and the full library, including the same distractors as the executor, but only selects skills; selected-skill recall thus tests whether the needed skills are recognizable without task execution.  In \emph{preselection-plus-execution}, the executor then sees only the selected subset; loaded-skill recall and task success measure the practical benefit.  The diagnosis changes only the objective, whereas the mitigation also narrows the executor's library.

\subsection{Metrics}
\emph{Pass@3} is task success, computed with the standard unbiased estimator over repeated runs.  \emph{Loaded-skill recall} is $|G\cap L|/|G|$, where $L$ is the set of skills opened during execution; the \emph{zero-load rate} is the percentage of runs with $|L|=0$, and \emph{average skills loaded} is the mean of $|L|$.  For explicit routing, \emph{selected-skill recall} is $|G\cap P|/|G|$, where $P$ is the preselected set.  Appendix~\ref{app:precision-metrics} reports precision, and Appendix~\ref{app:non-golden-audit} analyzes the loading of non-golden skills.

\section{Main Findings}
\label{sec:experiment}
\subsection{Experimental setup}
\label{sec:exp-setup}
\paragraph{SkillsBench tasks.}  SkillsBench \citep{li2026skillsbench} provides task instructions, workspace files, executable verifiers, and task-specific golden skills.  We exclude five tasks with infrastructure or environment problems and use the remaining 82 (Appendices~\ref{app:skillsbench-details} and~\ref{app:main-task-exclusions}).  Runs are costly: under implicit routing, one run averages 15.3 minutes and 485k input tokens without irrelevant skills, rising to 17.0 minutes and 1.95M input tokens with 1,000.

\paragraph{Skill libraries.}  We downloaded more than 20,000 skills from a skill market in leaderboard order and sample irrelevant skills from this pool with a fixed seed (42), so the distractors are identical across models and strategies.  We evaluate $s\in\{0,10,100,1000,10000\}$, always installing the golden skills.  Every preselector exposes at most five skills: LLM-based selectors choose up to five, and RAG returns the top five by embedding similarity.

\paragraph{Models and scale.}  Given this cost, the main experiments use three open-weight executors: GLM-5 \citep{glm2026glm5}, Kimi K2.5 \citep{kimi2026k25}, and MiniMax M2.5 \citep{minimax2026m25}; RAG and hybrid preselection use Qwen text-embedding-v4 embeddings.  Each model runs every task and condition three times, and we average over the task--model--run grid, for more than 15,000 trajectories and 10 billion tokens in total.  A budget-limited validation on Claude Opus 4.6, Gemini 3.1 Pro, and GPT-5.4 yields consistent conclusions (Appendix~\ref{app:proprietary-validation}).

\subsection{Lost in Implicit Routing}
\label{sec:lost-implicit-routing}
\begin{figure*}[t]
\centering
\includegraphics[width=\textwidth]{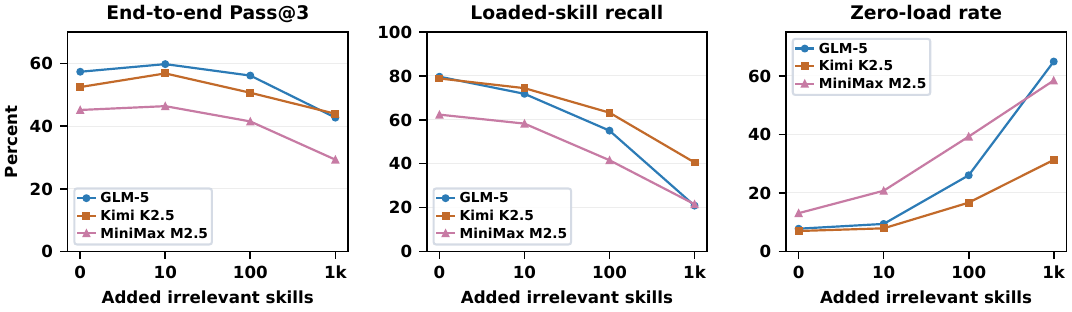}
\caption{Implicit-routing scaling by execution model.  (a) End-to-end Pass@3.  (b) Loaded-skill recall.  (c) Zero-load rate.  Only the number of added irrelevant skills changes.}
\label{fig:naive-model-scaling}
\end{figure*}

\subsubsection{Implicit loading degrades before task success}
Figure~\ref{fig:naive-model-scaling} shows implicit routing for the three open-weight models.  \textbf{Implicit routing collapses as irrelevant skills grow.}  With only 100 irrelevant skills, loaded-skill recall already falls by 24.6, 15.6, and 20.7 points for GLM-5, Kimi K2.5, and MiniMax M2.5 (Table~\ref{tab:model-loaded-selected}), while zero-load rates rise: agents increasingly stop using skills at all rather than merely choosing wrong ones.  Task success lags behind.  At the same scale, GLM-5 keeps Pass@3 almost unchanged (57.3\% to 56.1\%) and Kimi K2.5 stays near its baseline (52.4\% to 50.6\%).  Strong executors can solve around missing skills, so final success alone hides a failing skill protocol.  Three proprietary models on a 37-task subset show the same pressure: Claude Opus 4.6, for example, keeps Pass@3 at 62.2\% with 1,000 irrelevant skills (vs.\ 64.9\%), yet its loaded-skill recall falls from 81.3\% to 30.9\% (Appendix~\ref{app:proprietary-validation}).  The collapse is also not confined to multi-skill tasks: at 1,000 irrelevant skills, loaded-skill recall falls from 91.9\% to 38.0\% on single-skill tasks and from 58.8\% to 14.8\% on tasks with four or more golden skills (Appendix~\ref{app:skill-demand-breakdown}).  Appendix~\ref{app:implicit-model-curve-data} lists all plotted values.

\subsubsection{Library growth suppresses skill use}
Table~\ref{tab:loading-propensity-audit} shows that from zero to 1,000 distractors, the average number of loaded skills falls from 1.72 to 0.66 and the zero-load rate rises from 9.2\% to 50.8\%.  Could agents instead be replacing golden skills with other skills?  Non-golden loading is rare (2.32\% of evaluations), and only 0.95\% of evaluations load a non-golden skill whose embedding cosine similarity to a missed golden skill exceeds 0.8.  Such loads concentrate at $s=1000$ (7.19\% and 3.79\%, respectively), and the above-threshold cases are near-duplicates such as \texttt{pdf-reader} for \texttt{pdf} or \texttt{excel} for \texttt{xlsx}.  When agents do load skills, the loads are mostly correct: under implicit routing at 1,000 irrelevant skills, loaded-skill precision is still 88.6\%, 87.2\%, and 76.4\% for the three models (Appendix~\ref{app:precision-metrics}).  By construction, no distractor duplicates a golden skill.  Routing collapse therefore primarily reflects under-loading, not substitution (details in Appendix~\ref{app:non-golden-audit}).

\begin{table}[t]
\centering
\small
\setlength{\tabcolsep}{4pt}
\begin{tabular}{rrrr}
\toprule
$s$ & Avg. skills loaded & ZeroLoad & Load Rec. \\
\midrule
0 & 1.72 & 9.2 & 73.5 \\
10 & 1.54 & 13.0 & 67.8 \\
100 & 1.20 & 27.9 & 52.9 \\
1000 & 0.66 & 50.8 & 27.6 \\
\bottomrule
\end{tabular}
\caption{Skill-loading behavior as the number of irrelevant skills grows.  ZeroLoad and loaded-skill recall are percentages.  This analysis covers 86 tasks, a slightly broader set than the main study (Appendix~\ref{app:non-golden-audit}).}
\label{tab:loading-propensity-audit}
\end{table}

\subsubsection{The same skills are recognizable when selected explicitly}
Are the relevant skills intrinsically hard to recognize?  Table~\ref{tab:model-loaded-selected} compares implicit loaded-skill recall with routing-only selected-skill recall for the same models and scales.  \textbf{The same LLMs still identify the needed skills when selection is explicit.}  At 1,000 irrelevant skills, GLM-5 loads 20.9\% of golden skills but selects 82.5\%, and Kimi K2.5 loads 40.5\% but selects 80.7\%; MiniMax M2.5 shows a smaller but still large gap (21.5\% vs.\ 58.4\%).  The full metadata list thus often suffices for recognition when selection is the sole objective; the failure arises when selection is folded into reading files, planning commands, and completing the task.

\begin{table}[t]
\centering
\small
\setlength{\tabcolsep}{4pt}
\begin{tabular}{llrrrr}
\toprule
Model & Measure & 0 & 10 & 100 & 1k \\
\midrule
GLM-5 & Implicit loaded & 79.6 & 71.8 & 55.0 & 20.9 \\
GLM-5 & Explicit selected & 98.4 & 93.1 & 91.7 & 82.5 \\
\midrule
Kimi K2.5 & Implicit loaded & 78.8 & 74.4 & 63.2 & 40.5 \\
Kimi K2.5 & Explicit selected & 98.5 & 93.3 & 90.7 & 80.7 \\
\midrule
MiniMax M2.5 & Implicit loaded & 62.3 & 58.2 & 41.6 & 21.5 \\
MiniMax M2.5 & Explicit selected & 98.3 & 88.0 & 83.6 & 58.4 \\
\bottomrule
\end{tabular}
\caption{Implicit loaded-skill recall (\%) versus explicit LLM selected-skill recall (\%) by model and number of added irrelevant skills.}
\label{tab:model-loaded-selected}
\end{table}

\subsection{Analyzing Explicit Routing}
\label{sec:explicit-routing-mitigation}
Since the bottleneck lies in routing during execution, we next ask how much preselection helps when the executor sees only the selected subset.  We do not propose a new routing algorithm; we compare simple preselectors to show when this design helps and what it costs.

\subsubsection{Explicit routing across library sizes}
\begin{figure*}[t]
\centering
\includegraphics[width=\textwidth]{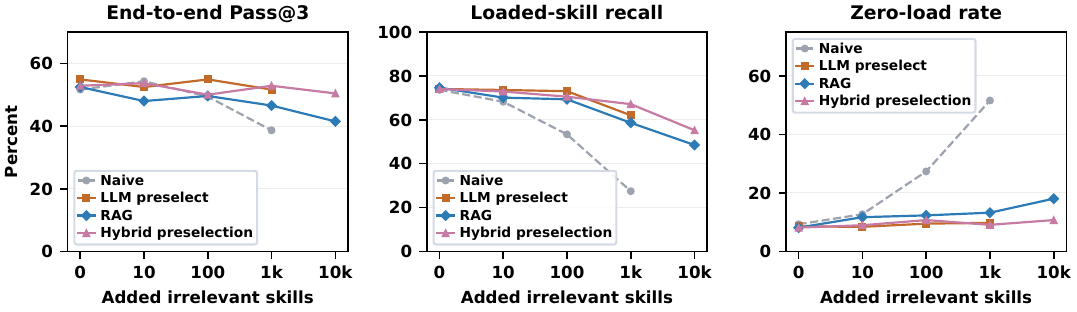}
\caption{Skill scaling curves by routing strategy.  (a) End-to-end Pass@3.  (b) Loaded-skill recall.  (c) Zero-load rate.}
\label{fig:skill-scaling}
\end{figure*}

Figure~\ref{fig:skill-scaling} compares implicit routing with the three preselectors.  \textbf{Narrowing the visible skill set reduces the scale effect}: Pass@3, loaded-skill recall, and zero-load rate all become less sensitive to library size.  The best strategy depends on scale.  LLM preselection is strongest around 100 added skills, whereas hybrid preselection is strongest at 1,000 and remains usable at 10,000, where full-list LLM preselection is omitted as infeasible.  At 100 added skills, LLM preselection raises aggregate loaded-skill recall from 53.3\% to 73.0\% and Pass@3 from 49.4\% to 54.9\%; at 10,000, hybrid preselection reaches 55.2\% recall and 50.4\% Pass@3, versus 48.5\% and 41.5\% for RAG (Appendix~\ref{app:strategy-curve-data}).  Table~\ref{tab:model-pass3} shows the same pattern for each executor: at 1,000 added skills, for instance, LLM preselection raises Pass@3 from 42.7\% to 58.5\% for GLM-5 and from 29.3\% to 42.7\% for MiniMax M2.5.

\begin{table}[t]
\centering
\small
\setlength{\tabcolsep}{3.5pt}
\begin{tabular}{lrrrrrr}
\toprule
Model & w/o & 0 & 10 & 100 & 1k & 10k \\
\midrule
\multicolumn{7}{l}{\textbf{Naive}} \\
GLM-5 & 26.8 & 57.3 & 59.8 & 56.1 & 42.7 & -- \\
Kimi K2.5 & 34.1 & 52.4 & 56.8 & 50.6 & 43.9 & -- \\
MiniMax M2.5 & 26.8 & 45.1 & 46.3 & 41.5 & 29.3 & -- \\
\midrule
\multicolumn{7}{l}{\textbf{LLM preselect}} \\
GLM-5 & 26.8 & 65.9 & 57.3 & 57.3 & 58.5 & -- \\
Kimi K2.5 & 34.1 & 54.9 & 56.1 & 58.5 & 53.7 & -- \\
MiniMax M2.5 & 26.8 & 43.9 & 43.9 & 48.8 & 42.7 & -- \\
\midrule
\multicolumn{7}{l}{\textbf{RAG}} \\
GLM-5 & 26.8 & 57.3 & 48.8 & 53.7 & 54.3 & 47.6 \\
Kimi K2.5 & 34.1 & 50.0 & 50.0 & 52.4 & 48.8 & 41.5 \\
MiniMax M2.5 & 26.8 & 50.0 & 45.1 & 42.7 & 36.6 & 35.4 \\
\midrule
\multicolumn{7}{l}{\textbf{Hybrid preselection}} \\
GLM-5 & 26.8 & 59.8 & 61.0 & 54.9 & 57.3 & 56.1 \\
Kimi K2.5 & 34.1 & 48.8 & 56.1 & 52.4 & 52.4 & 56.1 \\
MiniMax M2.5 & 26.8 & 50.0 & 43.9 & 42.7 & 48.8 & 39.0 \\
\bottomrule
\end{tabular}
\caption{Per-model Pass@3 (\%) by routing strategy and number of added irrelevant skills.  The w/o column disables skills entirely; ``--'' marks omitted infeasible settings.}
\label{tab:model-pass3}
\end{table}

Table~\ref{tab:selection-recall} reports selected-skill recall, which concerns the preselector's choice rather than whether the executor later opens the skills, and should be read with selected precision (Table~\ref{tab:selection-precision}).  RAG has the highest recall at 10 and 100 added skills, partly because it always returns five skills, but its precision is much lower (47.6\% and 43.9\%, vs.\ 99.8\% and 84.5\% for hybrid preselection).  At large scale, hybrid preselection is preferable: it has the highest recall at 1,000 and improves over RAG at 10,000 (62.0 to 72.2) while also raising precision from 30.1\% to 52.6\%.

\begin{table}[t]
\centering
\small
\setlength{\tabcolsep}{4pt}
\begin{tabular}{lrrrrr}
\toprule
Method & 0 & 10 & 100 & 1k & 10k \\
\midrule
LLM & 98.4 & 91.5 & 88.7 & 73.9 & -- \\
RAG & 98.6 & 95.9 & 90.0 & 74.1 & 62.0 \\
Hybrid & 98.2 & 91.4 & 87.9 & 81.9 & 72.2 \\
\bottomrule
\end{tabular}
\caption{Preselected-skill recall (\%) by method and number of added irrelevant skills.}
\label{tab:selection-recall}
\end{table}

\subsubsection{Token cost of routing}
Table~\ref{tab:token-decomp} separates LLM input and output tokens for execution and preselection; embedding cost is excluded because it is negligible (below \$1 in total).  Preselection adds a routing call but can sharply shrink the execution context in large libraries.  At 1,000 irrelevant skills, implicit routing consumes 1952k agent-input tokens, whereas LLM preselection spends 36k input tokens on routing and reduces agent input to 497k, for 547k in total.  RAG is cheapest at 10,000 skills (469k) because it makes no LLM call, but it also performs worst at that scale (Figure~\ref{fig:skill-scaling}).  Hybrid preselection costs modestly more (547k, including 8k for preselection) and generalizes better.

\begin{table}[t]
\centering
\scriptsize
\setlength{\tabcolsep}{2pt}
\begin{tabular}{@{}clrrrrr@{}}
\toprule
$s$ & Strat. & A-In & A-Out & P-In & P-Out & Total \\
\midrule
0 & Naive & 485k & 13k & 0k & 0k & 498k \\
0 & LLM & 481k & 13k & 0k & 0k & 494k \\
0 & RAG & 494k & 13k & 0k & 0k & 507k \\
0 & Hybrid & 499k & 13k & 0k & 0k & 513k \\
\midrule
10 & Naive & 481k & 13k & 0k & 0k & 494k \\
10 & LLM & 495k & 13k & 1k & 0k & 509k \\
10 & RAG & 481k & 13k & 0k & 0k & 494k \\
10 & Hybrid & 502k & 14k & 1k & 1k & 519k \\
\midrule
100 & Naive & 619k & 13k & 0k & 0k & 632k \\
100 & LLM & 490k & 13k & 4k & 1k & 508k \\
100 & RAG & 493k & 13k & 0k & 0k & 506k \\
100 & Hybrid & 476k & 13k & 6k & 1k & 497k \\
\midrule
1000 & Naive & 1952k & 13k & 0k & 0k & 1966k \\
1000 & LLM & 497k & 13k & 36k & 1k & 547k \\
1000 & RAG & 490k & 13k & 0k & 0k & 503k \\
1000 & Hybrid & 533k & 14k & 6k & 2k & 554k \\
\midrule
10000 & RAG & 456k & 13k & 0k & 0k & 469k \\
10000 & Hybrid & 525k & 14k & 6k & 2k & 547k \\
\bottomrule
\end{tabular}
\caption{Token decomposition by skill-pool size and strategy.  A/P denote execution-agent/preselection tokens; In/Out denote input/output tokens.}
\label{tab:token-decomp}
\end{table}

\subsubsection{Why hybrid preselection helps at 10k}
A single RAG query built from the task instruction can over-focus on one part of a multi-step task and miss other required capabilities.  Hybrid preselection instead has an LLM write several task-focused queries, retrieves candidates for each, and lets an LLM select from the pooled candidates.  Table~\ref{tab:sq-vs-mq} compares this pool with single-query RAG@100 before final selection: at 10,000 irrelevant skills, it raises candidate recall from 88.2 to 91.8 and all-gold coverage from 78.0 to 86.2, so needed skills are less often missing from the pool.  To merge the per-query lists, we take a fixed quota of top unseen skills from each query, which gives higher all-gold coverage at 10,000 than max-score or mean-score merging (86.2 vs.\ 80.5 and 78.2; Appendix~\ref{app:candidate-aggregation}).

\begin{table}[t]
\centering
\small
\setlength{\tabcolsep}{5pt}
\begin{tabular}{crrr}
\toprule
$s$ & SQ Rec. & Hybrid Rec. & $\Delta$ Rec. \\
\midrule
0 & 99.4 & 99.4 & +0.0 \\
10 & 99.4 & 99.4 & +0.0 \\
100 & 99.4 & 99.4 & +0.0 \\
1000 & 93.6 & 96.3 & +2.7 \\
10000 & 88.2 & 91.8 & +3.6 \\
\midrule
$s$ & SQ All & Hybrid All & $\Delta$ All \\
\midrule
0 & 98.8 & 98.9 & +0.1 \\
10 & 98.8 & 98.9 & +0.1 \\
100 & 98.8 & 98.9 & +0.1 \\
1000 & 87.8 & 92.0 & +4.1 \\
10000 & 78.0 & 86.2 & +8.2 \\
\bottomrule
\end{tabular}
\caption{Single-query RAG@100 versus the hybrid candidate pool.  Values are percentages.  Top: individual recall; bottom: all-gold coverage.}
\label{tab:sq-vs-mq}
\end{table}

\paragraph{Coverage and latency trade-offs.}  Fixed preselection can omit skills that become necessary later.  As distractors grow from 10 to 1,000, the average number of golden skills exposed to the executor falls from 2.21 to 1.78 for LLM preselection, from 2.38 to 1.85 for RAG, and from 2.21 to 1.97 for hybrid preselection, compared with 2.52 when all golden skills are visible (Appendix~\ref{app:preselection-tradeoffs}).  Coverage alone does not determine success: at 1,000, RAG exposes more golden skills than LLM preselection but reaches lower Pass@3 (46.5\% vs.\ 51.6\%), suggesting that which skills survive selection also matters.  Preselection also adds latency before execution even when it saves tokens: from 10 to 1,000 distractors, the median routing time rises from 8 to 19 seconds for LLM preselection and from 33 to 41 seconds for hybrid preselection, which makes two LLM calls (Table~\ref{tab:preselection-latency}).  RAG needs no LLM call; because we embedded all skills and task instructions in a batch before the experiments, its preselection reduces to a similarity lookup, so we do not report a separate latency for it.  Preselection is thus a useful mitigation rather than a complete solution, and its benefit must be weighed against coverage loss and response time.

\section{Why Agents Are Lost in Implicit Routing}
\label{sec:error-analysis}

\subsection{Experimental setup}
We use a 10-task subset at 1,000 irrelevant skills (Appendix~\ref{app:experiment-subset}) and the three open-weight models, and test two interventions, each run three times per model.  \textbf{Masked} keeps every distractor slot but replaces its name, description, and location with \texttt{<mask>} placeholders of approximately the original length.  \textbf{Plan-first prompt} keeps the metadata intact but instructs the agent to consider skills before execution.  Table~\ref{tab:implicit-routing-results} adds the no-extra-skills reference and preselection-plus-execution on the same subset.  Field-level variants mask only names, descriptions, locations, or names plus locations; each runs once per task and model and is compared with a single run of the original condition, with full masking as the three-run reference.  In every variant, golden-skill metadata stays readable (prompts in Appendix~\ref{app:intervention-prompts}).

\begin{table}[t]
\centering
\scriptsize
\setlength{\tabcolsep}{4pt}
\begin{tabular}{lrrr}
\toprule
Setting & Load Rec. & Zero & All-gold \\
\midrule
No extra skills & 73.5 & 5.6 & 52.2 \\
\midrule
\multicolumn{4}{l}{\textit{Implicit routing, 1,000 irrelevant skills}} \\
Original & 18.6 & 62.2 & 6.7 \\
Masked & 63.9 & 23.6 & 49.4 \\
Plan-first prompt & 44.6 & 15.7 & 25.8 \\
\midrule
\multicolumn{4}{l}{\textit{Explicit preselection, 1,000 irrelevant skills}} \\
LLM & 50.4 & 10.0 & 30.0 \\
RAG & 56.4 & 21.6 & 29.5 \\
Hybrid & 61.1 & 11.1 & 38.9 \\
\bottomrule
\end{tabular}
\caption{Routing results (\%) on the 10-task subset.  All rows except the first use 1,000 irrelevant skills.  Load Rec.\ is loaded-skill recall; Zero is the zero-load rate; All-gold is the share of runs that load or select every golden skill.}
\label{tab:implicit-routing-results}
\end{table}

\subsection{Results and discussion}
\textbf{Skill metadata creates semantic interference.}  In Table~\ref{tab:implicit-routing-results}, masking keeps the context long, yet largely restores no-extra-skills loading (63.9\% vs.\ 73.5\% recall), cuts the zero-load rate from 62.2\% to 23.6\%, and is competitive with the strongest preselection row (61.1\%).  Because the slots and approximate length are preserved, the gain is not an artifact of a shorter context.  Long metadata alone therefore does not cause routing collapse; the failure appears when distractors carry meaningful descriptions that look like plausible alternative actions.

\paragraph{Descriptions dominate the field-level effect.}  Masking only descriptions raises loaded-skill recall to 63.3\%, close to full masking (63.9\%; Table~\ref{tab:field-masking}).  Masking names or locations alone helps less (43.7\% and 47.5\%), and masking both together reaches only 51.7\%.  The latter removes the explicit targets for \texttt{load\_skill} and direct file reads, although agents could still find the files by exploring.  That the remaining descriptions still suppress golden-skill loading shows that their effect extends beyond competition among explicit loading targets.

\begin{table}[t]
\centering
\small
\setlength{\tabcolsep}{3pt}
\begin{tabular}{lrrr}
\toprule
Setting & Avg. loaded & ZeroLoad & Load Rec. \\
\midrule
Original metadata & 0.50 & 60.0 & 18.3 \\
Mask names & 0.93 & 40.0 & 43.7 \\
Mask descriptions & 1.17 & 23.3 & 63.3 \\
Mask locations & 0.90 & 36.7 & 47.5 \\
Mask names + locations & 1.10 & 36.7 & 51.7 \\
\midrule
Mask all distractor (reference) & 1.26 & 23.6 & 63.9 \\
\bottomrule
\end{tabular}
\caption{Field-level masking of distractor metadata on the 10-task subset.  Avg.\ loaded is the mean number of skills loaded; ZeroLoad and Load Rec.\ are percentages.  The first five rows use one run per task and model; full masking is the three-run reference.}
\label{tab:field-masking}
\end{table}

\textbf{Implicit routing suffers from instruction interference.}  The Plan-first prompt improves over the original condition but remains weaker than explicit preselection.  To isolate the initial routing decision, Table~\ref{tab:plan-vs-preselect} compares the skills planned in the agent's first response with isolated LLM preselection; this planned recall is lower than final loaded recall because agents sometimes load skills absent from the initial plan.  Both settings see nearly the same context, the task instruction plus the full metadata, and differ mainly in the system prompt: LLM preselection asks only for skill selection, whereas Plan-first asks the agent to solve the task and to choose skills first.  Despite this explicit reminder, Plan-first recalls far fewer golden skills (39.9\% vs.\ 70.6\%) and covers all golden skills in far fewer runs (20.5\% vs.\ 54.4\%).  Routing and beginning the task thus compete within the same response.

\begin{table}[t]
\centering
\small
\setlength{\tabcolsep}{5pt}
\begin{tabular}{lrr}
\toprule
Setting & Recall & All-gold \\
\midrule
Plan-first prompt & 39.9 & 20.5 \\
LLM preselection & 70.6 & 54.4 \\
\bottomrule
\end{tabular}
\caption{Golden-skill selection quality (\%) in the Plan-first response versus isolated LLM preselection on the same 10-task subset.  Recall is computed against the task's golden skills.}
\label{tab:plan-vs-preselect}
\end{table}

\paragraph{Descriptions have two functional roles.}  These results suggest a link between the two forms of interference.  During execution, descriptions act as conditional sub-instructions (load this skill when its situation arises), creating a multi-instruction-following problem alongside the task.  In routing-only selection, they are instead semantic-matching inputs rather than instructions to monitor.  This role shift helps explain why agents select skills more reliably when selection is their sole objective, even with the full library visible.

\subsection{Case study}
Traces further support instruction interference.  Figure~\ref{fig:nudge-case-boxes} shows two plan-first failures.  In Case~1, GLM-5 explicitly plans to load \emph{pid-controller}, \emph{vehicle-dynamics}, and \emph{simulation-metrics}, but opens only the first.  In Case~2, MiniMax M2.5 writes a step-by-step plan for a D3 visualization that mentions no skill, and never loads \emph{d3-visualization}.  Both runs fail.

\begin{figure}[t]
\centering
\scriptsize
\begin{tracecase}{Case 1: planned but not loaded}
\textbf{Task/model:} \emph{adaptive-cruise-control}, GLM-5.\\
\textbf{Agent plan excerpt:} ``Let me load the relevant skills first: 1. \textbf{\emph{pid-controller}} -- For implementing PID control loops; 2. \textbf{\emph{vehicle-dynamics}} -- For vehicle motion simulation and cruise control modes; 3. \textbf{\emph{simulation-metrics}} -- For calculating performance metrics.''\\
\textbf{Actually loaded:} \emph{pid-controller}.\\
\textbf{Missed:} \emph{vehicle-dynamics}, \emph{simulation-metrics}, \ldots.  The run fails.
\end{tracecase}
\vspace{1mm}
\begin{tracecase}{Case 2: no skill plan}
\textbf{Task/model:} \emph{data-to-d3}, MiniMax M2.5.\\
\textbf{Agent plan excerpt:} ``Explore the input data files; set up the output directory; copy the input data; download D3.js v6; create the visualization with bubble chart and data table; add interactive features.''\\
\textbf{Actually loaded:} none.\\
\textbf{Missed:} \emph{d3-visualization}.  The run fails.
\end{tracecase}
\caption{Two trace cases under the plan-first prompt.  Actually loaded counts both \texttt{load\_skill} calls and direct reads of skill files.}
\label{fig:nudge-case-boxes}
\end{figure}

A prompt nudge thus does not reliably separate routing from execution; reducing the visible skill set addresses both forms of interference more directly.

\section{Practical Recommendations}
Because irrelevant descriptions interfere with both routing and task execution, narrowing the visible skill set before execution is more reliable than strengthening prompt instructions, and the right method depends on scale.  For small curated libraries, implicit routing remains usable, but systems should track which skills are selected and loaded rather than rely on task success alone.  Around the hundred-skill scale, explicit preselection should become the default; direct LLM preselection is simple and strong while the full metadata list fits in context.  For thousands of skills, full-list LLM preselection becomes expensive or infeasible, and hybrid preselection (retrieval for a compact candidate set, then an LLM selector) is the strongest option in our experiments, improving relevant-skill coverage while keeping the execution context small.  Since fixed preselection can omit skills needed later and adds latency, deployments should also monitor coverage and response time.  Prompt reminders to consider skills are a lightweight safeguard that should complement, not replace, preselection.  For evaluation, benchmarks of skill-using agents should report loaded-skill recall and zero-load rate alongside task success, since Pass@3 alone can hide routing failure.

\section{Related Work}
\paragraph{LLM agents and tool choice.} ReAct couples reasoning with actions \citep{yao2023react}, Toolformer learns tool calls \citep{schick2023toolformer}, and MRKL combines language models with external knowledge and reasoning modules \citep{karpas2022mrkl}.  As action spaces grow, Gorilla and ToolLLM study API selection and invocation over massive API sets \citep{patil2023gorilla,qin2023toolllm}, MetaTool evaluates whether models know when and which tools to use \citep{huang2024metatool}, and RAG-MCP retrieves tools to counter prompt bloat \citep{gan2025ragmcp}.  Overlapping tool names and descriptions also cause selection ambiguity, motivating tool merging or filtering \citep{liu2025toolscope}.

\paragraph{Skills versus tools.} Skills are more general than callable tools: they may package scripts or command-line workflows, but may also contain only background knowledge, constraints, procedures, or examples \citep{anthropicEquippingSkills2025}.  They also enter context differently.  Tools are typically exposed through callable interfaces and argument schemas \citep{anthropicWritingTools2025}, whereas skills use progressive disclosure: the agent first sees compact metadata, and full instructions, script arguments, or implementation details become visible only after loading \citep{anthropicAgentSkills2025,anthropicSkills2026}.  The loading decision must therefore be made from metadata alone.

\paragraph{Reusable experience and skills.} Voyager writes and retrieves executable code skills in Minecraft \citep{wang2023voyager}, Reflexion stores verbal feedback \citep{shinn2023reflexion}, and ExpeL extracts natural-language lessons from prior experience \citep{zhao2023expel}.  More recently, ReUseIt synthesizes reusable web workflows \citep{liu2025reuseit}, SkillWeaver discovers and refines web-agent skills \citep{zheng2025skillweaver}, and SkillRL distills trajectories into hierarchical skill banks for reinforcement learning \citep{xia2026skillrl}.  These works focus on acquiring skills; we assume an existing library and ask whether agents can still route to the right skills as it grows.

\paragraph{Long-context use and retrieval.} Retrieval-augmented generation retrieves evidence before generation \citep{lewis2020retrieval}, dense passage retrieval enables scalable semantic search \citep{karpukhin2020dense}, and Fusion-in-Decoder studies how generators consume retrieved passages \citep{izacard2021leveraging}; HyDE and Self-RAG improve retrieval with hypothetical documents or adaptive retrieval decisions \citep{gao2023hyde,asai2024selfrag}.  Long-context and needle-in-a-haystack studies show that models do not always use relevant information placed in context \citep{liu2024lost,bai2024longbench,kamradt2023needle}, and \citet{du2025contextLength} find that context length alone can hurt performance even with perfect retrieval and masked irrelevant tokens.  Skill routing behaves differently: replacing irrelevant metadata with placeholders largely restores golden-skill loading, so the collapse stems mainly from semantic and instruction interference rather than length alone (Section~\ref{sec:error-analysis}).

\section{Conclusion}
We documented skill routing collapse: as libraries grow, agents increasingly load fewer or no skills even though the relevant skills remain available, and loading other skills instead is too rare to explain this decline.  Masking only irrelevant descriptions largely restores skill loading, whereas explicitly planning skill use with the metadata intact remains less effective than preselection.  These findings point to semantic interference from descriptions and instruction interference when skill choice competes with task completion.  Preselection can stabilize skill use but introduces coverage and latency trade-offs.  Growing skill libraries therefore call for explicit routing and for evaluating skill use, not task success alone.  A natural next step is routing that revisits the selected subset during execution, recovering skills that fixed preselection omits without exposing the full library.

\clearpage
\section*{Limitations}
For budget reasons, the proprietary-model validation covers only a 37-task subset and not every routing setting of the main study, so we treat it as a consistency check (Appendix~\ref{app:proprietary-validation}).

Field-level masking uses a small subset and fewer repetitions than full masking, so numerical proximity between conditions does not establish equivalence.  Name-plus-location masking hides identifiers in the prompt but does not delete skill files.  Embedding similarity screens plausible alternatives, not functional interchangeability.  The two roles we attribute to descriptions are an interpretation of behavioral evidence, not a measurement of internal model processing.  Routing latency depends on API serving conditions and does not establish end-to-end speedup; because RAG used embeddings computed in advance, our setup does not capture the cost of embedding new tasks online.  Fixed preselection can omit skills needed later (Appendix~\ref{app:preselection-tradeoffs}).

\bibliography{references}
\clearpage
\appendix
\raggedbottom
\setcounter{topnumber}{10}
\setcounter{bottomnumber}{10}
\setcounter{totalnumber}{20}
\setcounter{dbltopnumber}{10}
\renewcommand{\topfraction}{0.95}
\renewcommand{\bottomfraction}{0.95}
\renewcommand{\textfraction}{0.05}
\renewcommand{\floatpagefraction}{0.8}
\renewcommand{\dbltopfraction}{0.95}
\renewcommand{\dblfloatpagefraction}{0.8}
\section{SkillsBench Dataset Details}
\label{app:skillsbench-details}
SkillsBench is a benchmark for evaluating how agent skills affect LLM-agent task solving \citep{li2026skillsbench}.  The original paper describes a community-driven construction process in which 105 contributors submitted 322 candidate tasks, followed by automated checks and human review.  It states that the benchmark contains 86 tasks, excludes two GPU-dependent tasks from evaluation, and evaluates 84 tasks across 11 domains.  Each task is a self-contained module with a human-written instruction, a containerized environment, a reference solution, deterministic verifier scripts, and one or more curated golden skills.

We use the official SkillsBench GitHub repository, which contains 87 tasks; our counts follow the repository.  We exclude the five tasks in Appendix~\ref{app:main-task-exclusions} for infrastructure or environment reliability reasons, leaving 82 adopted tasks in the main controlled study.  Table~\ref{tab:adopted-skill-counts} reports how many golden skills these adopted tasks contain.  Most tasks require only a small number of skills, but the benchmark still includes multi-skill workflows that test whether routing can recover several complementary procedures.

\begin{table}[!htbp]
\centering
\small
\setlength{\tabcolsep}{10pt}
\begin{tabular}{cc}
\toprule
$|G|$ & Tasks \\
\midrule
1 & 26 \\
2 & 20 \\
3 & 16 \\
4 & 11 \\
5 & 6 \\
6 & 2 \\
7 & 1 \\
\bottomrule
\end{tabular}
\caption{Distribution of golden-skill counts among the 82 adopted SkillsBench tasks.  $|G|$ is the number of golden skills for a task.}
\label{tab:adopted-skill-counts}
\end{table}

Table~\ref{tab:implicit-run-resources} gives the average resource footprint of one valid implicit-routing run as the number of added irrelevant skills increases.  The table averages over the adopted tasks and the three open-weight execution models.  Even before adding irrelevant skills, a single run averages more than fifteen minutes and roughly half a million input tokens, which is why the full study is expensive and why the 1,000-skill diagnostic interventions use a smaller task subset.

\begin{table}[!htbp]
\centering
\small
\setlength{\tabcolsep}{6pt}
\begin{tabular}{rrrrr}
\toprule
$s$ & Runs & Time & Input toks & Output toks \\
\midrule
0 & 730 & 15.3m & 485k & 13k \\
10 & 731 & 16.7m & 481k & 13k \\
100 & 728 & 15.4m & 619k & 13k \\
1000 & 727 & 17.0m & 1.95M & 13k \\
\bottomrule
\end{tabular}
\caption{Average valid-run resource footprint under implicit routing by number of added irrelevant skills.  Time is wall-clock execution time per run; token columns are LLM input and output tokens per run.}
\label{tab:implicit-run-resources}
\end{table}

\subsection{License compliance}
The SkillsBench repository includes an Apache License 2.0 license file.  We do not modify or redistribute the benchmark; we use it as an evaluation benchmark and report the resulting experimental measurements.  We preserve attribution through citation to the original paper.  Our use is therefore consistent with the permissions and notice requirements of Apache License 2.0.

\subsection{Server configuration}
\label{app:server-config}
The main experiments were executed on Linux servers with an AMD EPYC 7K62 processor (16 physical cores, 32 hardware threads) and 62.3 GiB of memory.  The field-level masking experiments ran in containerized Linux environments on an Apple Silicon machine with 16 GB of memory; they measure skill-loading behavior and are not used for latency comparisons.  LLM inference was provided through Tencent TokenPlan.

\section{Main Study Task Exclusions}
\label{app:main-task-exclusions}
The main controlled study excludes the five tasks listed in Table~\ref{tab:main-task-exclusions} because we encountered infrastructure or environment problems that prevented reliable execution.

\begin{table}[!htbp]
\centering
\small
\begin{tabular}{rl}
\toprule
\# & Task \\
\midrule
1 & fix-visual-stability \\
2 & mhc-layer-impl \\
3 & multilingual-video-dubbing \\
4 & python-scala-translation \\
5 & speaker-diarization-subtitles \\
\bottomrule
\end{tabular}
\caption{Tasks excluded from the main controlled study for infrastructure or environment reasons.}
\label{tab:main-task-exclusions}
\end{table}

\section{Experimental Subset Tasks}
\label{app:experiment-subset}
For budget reasons, the controlled interventions in Section~\ref{sec:error-analysis} use the 10-task subset listed in Table~\ref{tab:experiment-subset-tasks}.  The same task list is used for the original, masked, plan-first, and explicit-preselection comparisons in that section.

\begin{table}[!htbp]
\centering
\small
\begin{tabular}{rl}
\toprule
\# & Task \\
\midrule
1 & adaptive-cruise-control \\
2 & data-to-d3 \\
3 & earthquake-phase-association \\
4 & econ-detrending-correlation \\
5 & enterprise-information-search \\
6 & manufacturing-equipment-maintenance \\
7 & react-performance-debugging \\
8 & reserves-at-risk-calc \\
9 & simpo-code-reproduction \\
10 & weighted-gdp-calc \\
\bottomrule
\end{tabular}
\caption{Tasks used in the 10-task experimental subset.}
\label{tab:experiment-subset-tasks}
\end{table}

\section{Full Data for Scaling Curves}
\label{app:scaling-curve-data}
The main text uses Figure~\ref{fig:naive-model-scaling} and Figure~\ref{fig:skill-scaling} to emphasize the shape of the scaling effects.  This appendix gives the exact values behind those curves so that the visual trends can be checked without reading points from the plots.  The full model-by-strategy Pass@3 matrix is already reported in Table~\ref{tab:model-pass3}; the additional tables here focus on the plotted slices and on the skill-loading measurements that are not otherwise visible in full.

\subsection{Implicit-routing scaling by execution model}
\label{app:implicit-model-curve-data}
Section~\ref{sec:lost-implicit-routing} argues that implicit routing breaks down before final task success fully reflects the failure.  Tables~\ref{tab:app-implicit-model-pass3}--\ref{tab:app-implicit-model-zero} provide the three panels of Figure~\ref{fig:naive-model-scaling} in tabular form.  Table~\ref{tab:app-implicit-model-pass3} reports end-to-end Pass@3 for the implicit-routing condition, Table~\ref{tab:app-implicit-model-loaded} reports the fraction of golden skills actually loaded during execution, and Table~\ref{tab:app-implicit-model-zero} reports the share of runs that load no skill at all.

The comparison across these tables shows the masking effect discussed in the main text.  Pass@3 declines for all three execution models by the 1,000-skill setting, but the routing metrics deteriorate earlier and more sharply.  For example, GLM-5 keeps Pass@3 near its small-library value at 100 irrelevant skills, while its loaded-skill recall has already dropped substantially and its zero-load rate has increased.  This is the empirical reason we treat loaded-skill recall and zero-load rate as first-class metrics rather than relying only on verifier success.

\begin{table}[!htbp]
\centering
\small
\setlength{\tabcolsep}{4pt}
\begin{tabular}{lrrrr}
\toprule
Model & 0 & 10 & 100 & 1k \\
\midrule
GLM-5 & 57.3 & 59.8 & 56.1 & 42.7 \\
Kimi K2.5 & 52.4 & 56.8 & 50.6 & 43.9 \\
MiniMax M2.5 & 45.1 & 46.3 & 41.5 & 29.3 \\
\bottomrule
\end{tabular}
\caption{Values for Figure~\ref{fig:naive-model-scaling}(a): implicit-routing Pass@3 (\%) by execution model and number of added irrelevant skills.}
\label{tab:app-implicit-model-pass3}
\end{table}

\begin{table}[!htbp]
\centering
\small
\setlength{\tabcolsep}{4pt}
\begin{tabular}{lrrrr}
\toprule
Model & 0 & 10 & 100 & 1k \\
\midrule
GLM-5 & 79.6 & 71.8 & 55.0 & 20.9 \\
Kimi K2.5 & 78.8 & 74.4 & 63.2 & 40.5 \\
MiniMax M2.5 & 62.3 & 58.2 & 41.6 & 21.5 \\
\bottomrule
\end{tabular}
\caption{Values for Figure~\ref{fig:naive-model-scaling}(b): implicit-routing loaded-skill recall (\%) by execution model.}
\label{tab:app-implicit-model-loaded}
\end{table}

\begin{table}[!htbp]
\centering
\small
\setlength{\tabcolsep}{4pt}
\begin{tabular}{lrrrr}
\toprule
Model & 0 & 10 & 100 & 1k \\
\midrule
GLM-5 & 7.8 & 9.4 & 26.0 & 64.9 \\
Kimi K2.5 & 7.0 & 7.9 & 16.7 & 31.3 \\
MiniMax M2.5 & 13.1 & 20.7 & 39.3 & 58.4 \\
\bottomrule
\end{tabular}
\caption{Values for Figure~\ref{fig:naive-model-scaling}(c): implicit-routing zero-load rate (\%) by execution model.  Lower is better.}
\label{tab:app-implicit-model-zero}
\end{table}

\subsection{Skill scaling by routing strategy}
\label{app:strategy-curve-data}
Section~\ref{sec:explicit-routing-mitigation} compares implicit routing with three explicit-routing strategies.  Tables~\ref{tab:app-strategy-pass3}--\ref{tab:app-strategy-zero} give the values behind Figure~\ref{fig:skill-scaling}: aggregate Pass@3, loaded-skill recall, and zero-load rate for each routing strategy and library size.  These tables are complementary to Table~\ref{tab:model-pass3}: the main-text table breaks task success down by execution model, whereas this appendix keeps the strategy-level view and adds the two skill-use metrics.

\begin{table}[!htbp]
\centering
\small
\setlength{\tabcolsep}{4pt}
\begin{tabular}{lrrrrr}
\toprule
Strategy & 0 & 10 & 100 & 1k & 10k \\
\midrule
Naive & 51.6 & 54.3 & 49.4 & 38.6 & -- \\
LLM preselect & 54.9 & 52.4 & 54.9 & 51.6 & -- \\
RAG & 52.4 & 48.0 & 49.6 & 46.5 & 41.5 \\
Hybrid preselection & 52.8 & 53.7 & 50.0 & 52.8 & 50.4 \\
\bottomrule
\end{tabular}
\caption{Values for Figure~\ref{fig:skill-scaling}(a): aggregate Pass@3 (\%) by routing strategy and number of added irrelevant skills.  ``--'' indicates omitted infeasible settings.}
\label{tab:app-strategy-pass3}
\end{table}

\begin{table}[!htbp]
\centering
\small
\setlength{\tabcolsep}{4pt}
\begin{tabular}{lrrrrr}
\toprule
Strategy & 0 & 10 & 100 & 1k & 10k \\
\midrule
Naive & 73.6 & 68.2 & 53.3 & 27.4 & -- \\
LLM preselect & 74.0 & 73.6 & 73.0 & 62.0 & -- \\
RAG & 74.6 & 70.1 & 69.3 & 58.6 & 48.5 \\
Hybrid preselection & 74.0 & 72.9 & 70.5 & 67.1 & 55.2 \\
\bottomrule
\end{tabular}
\caption{Values for Figure~\ref{fig:skill-scaling}(b): loaded-skill recall (\%) by routing strategy.}
\label{tab:app-strategy-loaded}
\end{table}

\begin{table}[!htbp]
\centering
\small
\setlength{\tabcolsep}{4pt}
\begin{tabular}{lrrrrr}
\toprule
Strategy & 0 & 10 & 100 & 1k & 10k \\
\midrule
Naive & 9.3 & 12.7 & 27.3 & 51.6 & -- \\
LLM preselect & 8.5 & 8.4 & 9.5 & 9.8 & -- \\
RAG & 8.1 & 11.7 & 12.3 & 13.2 & 18.0 \\
Hybrid preselection & 8.1 & 8.9 & 10.7 & 9.1 & 10.7 \\
\bottomrule
\end{tabular}
\caption{Values for Figure~\ref{fig:skill-scaling}(c): zero-load rate (\%) by routing strategy.  Lower is better.}
\label{tab:app-strategy-zero}
\end{table}

\section{Budget-Limited Proprietary Model Validation}
\label{app:proprietary-validation}
The main experiments use open-weight models to keep evaluation cost manageable.  This appendix asks whether the same implicit-routing pressure is visible for stronger proprietary frontier models.  Because these runs are substantially more expensive, we evaluate a smaller 37-task subset and treat the result as a qualitative consistency check rather than a second full benchmark.

Table~\ref{tab:proprietary-task-list} lists the exact proprietary-validation subset.

\begin{table}[!htbp]
\centering
\scriptsize
\setlength{\tabcolsep}{6pt}
\begin{tabular}{ll}
\toprule
Task & Task \\
\midrule
3d-scan-calc & flink-query \\
adaptive-cruise-control & flood-risk-analysis \\
azure-bgp-oscillation-route-leak & gh-repo-analytics \\
citation-check & glm-lake-mendota \\
dynamic-object-aware-egomotion & gravitational-wave-detection \\
earthquake-phase-association & grid-dispatch-operator \\
earthquake-plate-calculation & hvac-control \\
econ-detrending-correlation & invoice-fraud-detection \\
energy-ac-optimal-power-flow & jax-computing-basics \\
energy-market-pricing & jpg-ocr-stat \\
enterprise-information-search & lab-unit-harmonization \\
exceltable-in-ppt & lake-warming-attribution \\
exoplanet-detection-period & latex-formula-extraction \\
financial-modeling-qa & lean4-proof \\
find-topk-similiar-chemicals & manufacturing-codebook-normalization \\
fix-build-agentops & manufacturing-equipment-maintenance \\
fix-build-google-auto & manufacturing-fjsp-optimization \\
fix-druid-loophole-cve & mario-coin-counting \\
fix-erlang-ssh-cve & \\
\bottomrule
\end{tabular}
\caption{Tasks used in the proprietary-model validation subset.}
\label{tab:proprietary-task-list}
\end{table}

Table~\ref{tab:proprietary-validation} reports Pass@3 on that shared subset for three proprietary models and, for comparison, the three open-weight models evaluated on the same tasks. Table~\ref{tab:proprietary-loaded-recall} then reports loaded-skill recall for the proprietary models.

The proprietary models show the same qualitative pressure from larger skill libraries.  Their task success is generally more stable than the open-weight models on this subset: Claude Opus 4.6 remains near 60 Pass@3 even at 1,000 irrelevant skills. 

However, Table~\ref{tab:proprietary-loaded-recall} makes the routing degradation clearer.  Claude Opus 4.6 drops from 81.3 loaded-skill recall with no irrelevant skills to 30.9 at 1,000; Gemini 3.1 Pro drops from 77.4 to 43.0; and GPT-5.4 drops from 76.2 to 49.3.  The decline is sometimes slower than in the open-weight main study, especially for Gemini and GPT, but the direction is the same.  Relevant skills remain installed, yet implicit execution-time routing becomes less reliable as the library grows.  This supports the conclusion of the main experiment: stronger final task performance does not imply that implicit routing remains healthy.

\begin{table}[!htbp]
\centering
\small
\setlength{\tabcolsep}{4pt}
\begin{tabular}{lrrrr}
\toprule
Model & 0 & 10 & 100 & 1k \\
\midrule
Claude Opus 4.6\textsuperscript{\ensuremath{\dagger}} & 64.9 & 62.2 & 59.5 & 62.2 \\
Gemini 3.1 Pro\textsuperscript{\ensuremath{\dagger}} & 59.5 & 51.4 & 48.6 & 51.4 \\
GPT-5.4\textsuperscript{\ensuremath{\dagger}} & 54.1 & 56.8 & 43.2 & 43.2 \\
GLM-5 & 54.1 & 62.2 & 54.1 & 35.1 \\
Kimi K2.5 & 48.6 & 58.3 & 50.0 & 45.9 \\
MiniMax M2.5 & 35.1 & 45.9 & 37.8 & 21.6 \\
\bottomrule
\end{tabular}
\caption{Pass@3 (\%) on the 37-task proprietary-model validation subset.  Numeric columns give the number of added irrelevant skills on top of each task's golden skills.  Models marked with \textsuperscript{\ensuremath{\dagger}} are proprietary; the remaining rows are the three open-weight execution models evaluated on the same task subset under implicit routing.}
\label{tab:proprietary-validation}
\end{table}

\begin{table}[!htbp]
\centering
\small
\setlength{\tabcolsep}{4pt}
\begin{tabular}{lrrrr}
\toprule
Model & 0 & 10 & 100 & 1k \\
\midrule
Claude Opus 4.6\textsuperscript{\ensuremath{\dagger}} & 81.3 & 77.7 & 43.3 & 30.9 \\
Gemini 3.1 Pro\textsuperscript{\ensuremath{\dagger}} & 77.4 & 70.4 & 57.9 & 43.0 \\
GPT-5.4\textsuperscript{\ensuremath{\dagger}} & 76.2 & 62.9 & 51.8 & 49.3 \\
\bottomrule
\end{tabular}
\caption{Loaded-skill recall (\%) on the 37-task proprietary-model validation subset.  The statistic counts golden skills whose full instructions are accessed during execution.}
\label{tab:proprietary-loaded-recall}
\end{table}

\section{Precision Metrics}
\label{app:precision-metrics}
This appendix reports precision metrics for runtime skill loading and preselection.  Tables~\ref{tab:open-loaded-precision} and \ref{tab:proprietary-loaded-precision} report loaded-skill precision for open-weight and proprietary models, respectively; Table~\ref{tab:selection-precision} reports preselected-skill precision.  As the number of skills grows, agents often load fewer skills or no skills at all, so loaded precision usually declines less sharply than recall.  Loaded precision is computed over runs with at least one loaded skill; selected precision is computed over nonempty preselected sets.

\begin{table}[!htbp]
\centering
\small
\setlength{\tabcolsep}{4pt}
\begin{tabular}{lrrrrr}
\toprule
Model & 0 & 10 & 100 & 1k & 10k \\
\midrule
\multicolumn{6}{l}{\textbf{Naive}} \\
GLM-5 & 99.6 & 99.5 & 97.1 & 88.6 & -- \\
Kimi K2.5 & 99.6 & 99.0 & 97.7 & 87.2 & -- \\
MiniMax M2.5 & 99.7 & 99.1 & 95.2 & 76.4 & -- \\
\midrule
\multicolumn{6}{l}{\textbf{LLM preselect}} \\
GLM-5 & 99.6 & 99.6 & 97.7 & 90.8 & -- \\
Kimi K2.5 & 99.5 & 99.4 & 96.6 & 84.9 & -- \\
MiniMax M2.5 & 99.7 & 99.5 & 97.2 & 78.6 & -- \\
\midrule
\multicolumn{6}{l}{\textbf{RAG}} \\
GLM-5 & 99.7 & 99.2 & 99.2 & 91.9 & 83.6 \\
Kimi K2.5 & 99.7 & 99.2 & 99.5 & 91.0 & 80.7 \\
MiniMax M2.5 & 99.7 & 98.7 & 98.8 & 90.2 & 75.3 \\
\midrule
\multicolumn{6}{l}{\textbf{Hybrid preselection}} \\
GLM-5 & 99.4 & 99.3 & 97.7 & 90.7 & 82.6 \\
Kimi K2.5 & 99.6 & 99.6 & 98.3 & 92.0 & 76.8 \\
MiniMax M2.5 & 99.3 & 99.6 & 97.2 & 86.1 & 70.3 \\
\bottomrule
\end{tabular}
\caption{Runtime loaded-skill precision (\%) for open-weight models by routing strategy and number of added irrelevant skills.}
\label{tab:open-loaded-precision}
\end{table}

\begin{table}[!htbp]
\centering
\small
\setlength{\tabcolsep}{4pt}
\begin{tabular}{lrrrr}
\toprule
Model & 0 & 10 & 100 & 1k \\
\midrule
Claude Opus 4.6\textsuperscript{\ensuremath{\dagger}} & 100.0 & 100.0 & 100.0 & 95.2 \\
Gemini 3.1 Pro\textsuperscript{\ensuremath{\dagger}} & 100.0 & 100.0 & 100.0 & 100.0 \\
GPT-5.4\textsuperscript{\ensuremath{\dagger}} & 100.0 & 97.3 & 90.5 & 82.7 \\
\bottomrule
\end{tabular}
\caption{Runtime loaded-skill precision (\%) for proprietary models under implicit routing on the validation subset.}
\label{tab:proprietary-loaded-precision}
\end{table}

\begin{table}[!htbp]
\centering
\small
\setlength{\tabcolsep}{4pt}
\begin{tabular}{lrrrrr}
\toprule
Method & 0 & 10 & 100 & 1k & 10k \\
\midrule
LLM preselect & 100.0 & 99.7 & 80.0 & 63.0 & -- \\
RAG & 100.0 & 47.6 & 43.9 & 37.1 & 30.1 \\
Hybrid preselection & 100.0 & 99.8 & 84.5 & 72.0 & 52.6 \\
\bottomrule
\end{tabular}
\caption{Preselected-skill precision (\%) by method and number of added irrelevant skills.}
\label{tab:selection-precision}
\end{table}

\section{Skill Demand Breakdown}
\label{app:skill-demand-breakdown}
The main results aggregate over tasks with different numbers of golden skills.  This section checks whether the scaling story is driven only by tasks with many required skills or whether it also appears for simpler tasks.  Table~\ref{tab:golden-bins} groups tasks by the number of golden skills in the task environment: single-skill tasks, tasks with two or three golden skills, and tasks with four or more golden skills.  Within each library size $s$, the table reports both Pass@3 and loaded-skill recall for Naive exposure, LLM preselection, RAG, and hybrid preselection.

The table supports the same interpretation as the main text while adding a task-demand perspective.  At 1,000 irrelevant skills, Naive exposure drops for both single-skill and multi-skill groups, showing that routing collapse is not limited to the hardest tasks.  Loaded-skill recall shows the mechanism more directly: with 1,000 irrelevant skills, Naive exposure loads far fewer golden skills than the explicit-routing strategies across all three demand bins.  Multi-skill tasks nevertheless make the mechanism more consequential: partial retrieval can find one useful skill but still miss another workflow needed for a passing solution.  This is why the paper emphasizes all-gold coverage, loaded-skill recall, and trace analysis in addition to Pass@3.

\begin{table*}[!t]
\centering
\scriptsize
\setlength{\tabcolsep}{3pt}
\begin{tabular}{ccr*{4}{rr}}
\toprule
$s$ & $|G|$ & Tasks & \multicolumn{2}{c}{Naive} & \multicolumn{2}{c}{LLM} & \multicolumn{2}{c}{RAG} & \multicolumn{2}{c}{Hybrid} \\
\cmidrule(lr){4-5}\cmidrule(lr){6-7}\cmidrule(lr){8-9}\cmidrule(lr){10-11}
 & & & P3 & Load & P3 & Load & P3 & Load & P3 & Load \\
\midrule
0 & 1 & 26 & 53.8 & 91.9 & 52.6 & 93.6 & 52.6 & 95.7 & 56.4 & 95.7 \\
 & 2-3 & 36 & 52.8 & 68.7 & 55.6 & 68.1 & 55.6 & 68.7 & 50.0 & 67.2 \\
 & $\ge4$ & 20 & 46.7 & 58.8 & 56.7 & 59.3 & 46.7 & 57.8 & 53.3 & 58.0 \\
\midrule
10 & 1 & 26 & 51.9 & 91.8 & 53.8 & 94.9 & 48.7 & 89.3 & 57.7 & 94.9 \\
 & 2-3 & 36 & 58.3 & 60.9 & 52.8 & 67.6 & 49.1 & 63.3 & 53.7 & 66.6 \\
 & $\ge4$ & 20 & 50.0 & 51.2 & 50.0 & 56.6 & 45.0 & 57.3 & 48.3 & 55.6 \\
\midrule
100 & 1 & 26 & 51.9 & 73.2 & 55.1 & 94.4 & 50.0 & 88.9 & 55.1 & 92.3 \\
 & 2-3 & 36 & 49.1 & 47.5 & 57.4 & 65.7 & 51.9 & 61.7 & 45.4 & 63.1 \\
 & $\ge4$ & 20 & 46.7 & 38.4 & 50.0 & 58.3 & 45.0 & 57.5 & 51.7 & 55.6 \\
\midrule
1000 & 1 & 26 & 38.5 & 38.0 & 52.6 & 79.5 & 42.3 & 72.0 & 55.1 & 89.3 \\
 & 2-3 & 36 & 37.0 & 26.8 & 51.9 & 54.8 & 47.2 & 52.8 & 50.9 & 59.0 \\
 & $\ge4$ & 20 & 41.7 & 14.8 & 50.0 & 52.4 & 50.9 & 51.5 & 53.3 & 53.0 \\
\midrule
10000 & 1 & 26 & -- & -- & -- & -- & 38.5 & 64.1 & 47.4 & 70.1 \\
 & 2-3 & 36 & -- & -- & -- & -- & 43.5 & 42.4 & 54.6 & 48.3 \\
 & $\ge4$ & 20 & -- & -- & -- & -- & 41.7 & 39.1 & 46.7 & 48.4 \\
\bottomrule
\end{tabular}
\caption{Pass@3 and loaded-skill recall (\%) by skill-pool size and number of golden skills.  The row group $s$ gives the number of added irrelevant skills; $|G|$ bins tasks by their number of golden skills.  Load is loaded-skill recall.}
\label{tab:golden-bins}
\end{table*}

\section{Hybrid Candidate Aggregation Details}
\label{app:candidate-aggregation}
Using multiple queries raises a design question: how should the retrieved lists be merged into one candidate pool?  Table~\ref{tab:candidate-agg} compares three intuitive aggregators.  \texttt{Max} keeps the best similarity score that any query gives a skill.  \texttt{Mean} averages scores across queries.  \texttt{Quota} takes a fixed number of top unseen skills from each query, encouraging coverage of different sub-procedures.  Quota is the default in this paper, and Algorithm~\ref{alg:hybrid-preselection} in Appendix~\ref{app:algorithms} gives the corresponding procedure.  Quota gives the strongest all-gold coverage at 10,000 irrelevant skills, which is important for multi-skill tasks where missing one required workflow can break execution.

\begin{table}[!htbp]
\centering
\small
\setlength{\tabcolsep}{5pt}
\begin{tabular}{clrrr}
\toprule
$s$ & Agg. & Rec & All & Any \\
\midrule
1000 & Max & 97.9 & 95.4 & 100.0 \\
1000 & Mean & 93.9 & 89.7 & 97.7 \\
1000 & Quota & 96.3 & 92.0 & 98.9 \\
\midrule
10000 & Max & 88.7 & 80.5 & 95.4 \\
10000 & Mean & 88.7 & 78.2 & 95.4 \\
10000 & Quota & 91.8 & 86.2 & 97.7 \\
\bottomrule
\end{tabular}
\caption{Hybrid candidate quality (\%) at budget 100.}
\label{tab:candidate-agg}
\end{table}

\section{Non-Golden Skill Loading}
\label{app:non-golden-audit}
\paragraph{Experimental scope.}
We examine skill loading for GLM-5, Kimi K2.5, and MiniMax M2.5 at $s\in\{0,10,100,1000\}$ across all three runs.  This analysis covers 86 tasks, whereas the main task-success study uses the adopted 82-task subset.  Table~\ref{tab:loading-propensity-audit} therefore summarizes a slightly broader task set than the main-study curves.

A skill counts as loaded if the agent calls \texttt{load\_skill} on it or reads its \texttt{SKILL.md} file; references to skills outside the exposed library do not count.  Each task's golden set is restricted to golden skills present in the library.

\paragraph{Similarity matching.}
For each evaluation, let $N=L\setminus G$ be the loaded non-golden skills and $M=G\setminus L$ the missed golden skills.  We compare every pair in $N\times M$, using the RAG embedding model on the full skill name and description, formatted as \texttt{name: description}.  An evaluation counts as loading a potential alternative when any pair has cosine similarity strictly greater than 0.8.  We neither match across tasks nor compare against golden skills already loaded.  The threshold screens semantic overlap; it does not certify functional equivalence.

\begin{table}[t]
\centering
\small
\setlength{\tabcolsep}{5pt}
\begin{tabular}{rrr}
\toprule
$s$ & Non-golden load rate & $>0.8$ alternative rate \\
\midrule
0 & 0.00 & 0.00 \\
10 & 0.26 & 0.00 \\
100 & 1.83 & 0.00 \\
1000 & 7.19 & 3.79 \\
\bottomrule
\end{tabular}
\caption{Non-golden skill loading (\%).  Non-golden load rate measures whether any non-golden skill is loaded; alternative rate additionally requires similarity above 0.8 to a missed golden skill in the same evaluation.}
\label{tab:non-golden-rates}
\end{table}

\paragraph{Alternative loading is uncommon.}
Non-golden loads occur in 2.32\% of evaluations across scales, and potential alternatives above the threshold occur in 0.95\%.  They concentrate at $s=1000$ (Table~\ref{tab:non-golden-rates}) and include clear semantic neighbors such as \texttt{pdf-reader} and \texttt{pdf}, or \texttt{excel} and \texttt{xlsx}.  Table~\ref{tab:non-golden-pairs} lists all above-threshold task--model--skill-pair groups, combining repeated runs.  Although non-golden loading increases at the largest scale, it remains far less common than the decline in skill use.  Thus, routing collapse primarily reflects under-loading, not substitution of other skills for golden skills.

\paragraph{Exact duplicates.}
Distractors are sampled so that none shares its identifier with a golden skill of the task; exact duplicates of golden skills therefore never appear as distractors.  Differently named near-duplicates can remain and are covered by the similarity analysis above.

\begin{table*}[t]
\centering
\small
\setlength{\tabcolsep}{5pt}
\begin{tabular}{llllr}
\toprule
Task & Model & Runs & Loaded non-golden $\rightarrow$ missed golden & Cosine \\
\midrule
court-form-filling & Kimi K2.5 & 2 & pdf-generator $\rightarrow$ pdf & 0.922 \\
financial-modeling-qa & Kimi K2.5 & 1, 2, 3 & pdf-reader $\rightarrow$ pdf & 0.903 \\
financial-modeling-qa & MiniMax M2.5 & 1, 2 & pdf-reader $\rightarrow$ pdf & 0.903 \\
invoice-fraud-detection & MiniMax M2.5 & 1, 3 & pdf-reader $\rightarrow$ pdf & 0.903 \\
latex-formula-extraction & GLM-5 & 3 & pdf-reader $\rightarrow$ pdf & 0.903 \\
latex-formula-extraction & MiniMax M2.5 & 1 & pdf-reader $\rightarrow$ marker & 0.826 \\
latex-formula-extraction & MiniMax M2.5 & 1 & pdf-reader $\rightarrow$ pdf & 0.903 \\
pdf-excel-diff & GLM-5 & 1 & pdf-reader $\rightarrow$ pdf & 0.903 \\
pdf-excel-diff & Kimi K2.5 & 2 & pdf-reader $\rightarrow$ pdf & 0.903 \\
pdf-excel-diff & MiniMax M2.5 & 2 & pdf-reader $\rightarrow$ pdf & 0.903 \\
pg-essay-to-audiobook & GLM-5 & 2, 3 & tts $\rightarrow$ gtts & 0.811 \\
protein-expression-analysis & Kimi K2.5 & 1 & excel $\rightarrow$ xlsx & 0.906 \\
reserves-at-risk-calc & Kimi K2.5 & 1 & excel $\rightarrow$ xlsx & 0.906 \\
sales-pivot-analysis & Kimi K2.5 & 2 & excel $\rightarrow$ xlsx & 0.887 \\
sales-pivot-analysis & Kimi K2.5 & 3 & pdf-reader $\rightarrow$ pdf & 0.888 \\
sales-pivot-analysis & MiniMax M2.5 & 3 & excel $\rightarrow$ xlsx & 0.887 \\
shock-analysis-demand & Kimi K2.5 & 1 & excel $\rightarrow$ xlsx & 0.906 \\
shock-analysis-demand & MiniMax M2.5 & 2, 3 & excel $\rightarrow$ xlsx & 0.906 \\
shock-analysis-supply & Kimi K2.5 & 2 & excel $\rightarrow$ xlsx & 0.906 \\
shock-analysis-supply & MiniMax M2.5 & 2 & excel $\rightarrow$ xlsx & 0.906 \\
video-tutorial-indexer & GLM-5 & 1 & transcribee $\rightarrow$ speech-to-text & 0.843 \\
weighted-gdp-calc & Kimi K2.5 & 1, 2, 3 & excel $\rightarrow$ xlsx & 0.906 \\
\bottomrule
\end{tabular}
\caption{All above-threshold skill-pair groups for the three open-weight models, all at $s=1000$.  Runs lists which of the three repeated runs contain the pair.  Similarities use full name-and-description embeddings; the same skill name can have task-specific descriptions.}
\label{tab:non-golden-pairs}
\end{table*}

\section{Preselection Trade-offs and Latency}
\label{app:preselection-tradeoffs}
\paragraph{Fixed coverage.}
Preselection reduces the executor's visible library but can remove a skill that becomes relevant later.  Table~\ref{tab:preselection-coverage} reports the average number of golden skills exposed to the executor, computed from the intersection of the selected set with task golden skills and averaged over task--model identities.  Task success uses the main-study Pass@3 values.  Each preselection method retains fewer golden skills as distractors grow.  Reduced coverage limits task success because omitted skills are unavailable during execution, although success also depends on which skills are missed and the quality of the remaining set.  For example, RAG can retain more golden skills than LLM preselection yet yield lower task success; aggregate coverage alone does not determine whether the most consequential skills survive selection.

\begin{table}[t]
\centering
\small
\setlength{\tabcolsep}{5pt}
\begin{tabular}{lrr}
\toprule
Setting & Avg. golden exposed & Pass@3 (\%) \\
\midrule
All golden, no preselection & 2.52 & 51.6 \\
Naive-10 & 2.52 & 54.3 \\
\midrule
RAG-10 & 2.38 & 48.0 \\
LLM-10 & 2.21 & 52.4 \\
Hybrid-10 & 2.21 & 53.7 \\
RAG-100 & 2.21 & 49.6 \\
LLM-100 & 2.11 & 54.9 \\
Hybrid-100 & 2.10 & 50.0 \\
RAG-1000 & 1.85 & 46.5 \\
LLM-1000 & 1.78 & 51.6 \\
Hybrid-1000 & 1.97 & 52.8 \\
\bottomrule
\end{tabular}
\caption{Fixed preselection trades reduced interference for missed golden-skill coverage.  Method suffixes denote the number of added irrelevant skills.}
\label{tab:preselection-coverage}
\end{table}

\paragraph{Routing-stage latency.}
We measure routing-stage latency on the adopted main-study tasks with the three open-weight models across all three runs.  Routing-stage latency is the time from the start of preselection until the selected metadata is passed to the executor.  For hybrid preselection, it covers query generation, query embedding and retrieval, and final LLM selection, but not the one-time construction of the skill embedding index.  Environment setup, task execution, and verification are excluded.

\begin{table}[t]
\centering
\small
\setlength{\tabcolsep}{5pt}
\begin{tabular}{lrrrr}
\toprule
Method & $s$ & Mean (s) & Median (s) & P90 (s) \\
\midrule
LLM & 10 & 11.7 & 8.0 & 26 \\
LLM & 100 & 18.0 & 9.5 & 45 \\
LLM & 1000 & 35.4 & 19.0 & 73 \\
\midrule
Hybrid & 10 & 38.9 & 33.0 & 75 \\
Hybrid & 100 & 43.7 & 37.0 & 89 \\
Hybrid & 1000 & 48.3 & 41.0 & 93 \\
\bottomrule
\end{tabular}
\caption{Observed routing-stage latency before task execution.  P90 is the 90th percentile.  The two methods were measured in separate experiment periods, so differences also reflect serving conditions.}
\label{tab:preselection-latency}
\end{table}

The median delay increases from 8 to 19 seconds for LLM preselection and from 33 to 41 seconds for Hybrid as distractors grow from 10 to 1,000.  Hybrid preselection makes two LLM calls (query generation and final selection), whereas LLM preselection makes one.  Because the two methods were run at different times, these numbers reflect the API serving conditions of each period and are not a controlled latency comparison.

We do not report routing latency for RAG.  It makes no generative selector call, and we embedded all skill metadata and task instructions in a single batch before running the experiments, so RAG preselection at run time reduces to a similarity lookup over precomputed vectors.  Embedding calls are typically fast relative to LLM generation, but in a deployment that embeds each new task online, this step would add a small cost that our setup does not capture.  Likewise, token savings need not translate into shorter end-to-end wall time, which also depends on task behavior, API latency, and environment operations.  Practical deployments should weigh coverage and task success against the added routing delay, especially for interactive tasks.

\section{Execution and Routing Algorithms}
\label{app:algorithms}
The main text summarizes the execution protocol; this appendix spells out the steps.  Algorithm~\ref{alg:skill-execution} receives a task, a skill set, a preselector, and an execution model.  Line 1 applies the preselector to decide which skills are exposed.  Line 2 formats the exposed skill metadata, and Line 3 builds the initial model context.  Lines 4--7 repeat model tool calls and append tool outputs to the context.  Lines 8--9 run the verifier and return its result.

Algorithm~\ref{alg:hybrid-preselection} gives the hybrid preselection procedure.  Line 1 generates multiple search queries from the task instruction.  Line 2 embeds the skill metadata, and Line 3 initializes an empty candidate set.  Lines 4--8 retrieve top unseen candidates for each generated query and merge them into one candidate pool.  Lines 9--10 ask the final selector to choose from that candidate pool and return the selected skills.

\begin{algorithm}[H]
\footnotesize
\caption{Skill-conditioned agent execution}
\label{alg:skill-execution}
\begin{algorithmic}[1]
\REQUIRE task $\tau=(I,F,V,G)$, skills $S_s$, preselector $\pi$, execution model $M_e$
\ENSURE verifier result $r$
\STATE $P \gets \pi(I,S_s)$
\STATE $M \gets \textnormal{FormatMetadata}(P)$
\STATE $x \gets \textnormal{BuildContext}(I,M)$
\WHILE{$\textnormal{HasToolCall}(y \gets M_e(x))$}
    \STATE $o \gets \textnormal{Execute}(y,F)$
    \STATE $x \gets \textnormal{Append}(x,y,o)$
\ENDWHILE
\STATE $r \gets V(F)$
\STATE \textbf{return} $r$
\end{algorithmic}
\end{algorithm}

\begin{algorithm}[H]
\footnotesize
\caption{Hybrid preselection}
\label{alg:hybrid-preselection}
\begin{algorithmic}[1]
\REQUIRE instruction $I$, skills $S_s$, models $M_p,M_{emb}$, queries $n$, candidate budget $B$, final budget $k$
\ENSURE selected skills $P$
\STATE $Q \gets M_p.\textnormal{GenerateQueries}(I,n)$
\STATE $E \gets \{(u, M_{emb}(m(u))) : u\in S_s\}$
\STATE $C \gets \emptyset$
\FOR{$q \in Q$}
    \STATE $v_q \gets M_{emb}(q)$
    \STATE $R_q \gets \textnormal{RankBySim}(v_q,E)$
    \STATE $C \gets C \cup \textnormal{TopUnseen}(R_q, \lceil B/n\rceil)$
\ENDFOR
\STATE $P \gets M_p.\textnormal{SelectSkills}(I,C,k)$
\STATE \textbf{return} $P$
\end{algorithmic}
\end{algorithm}

Algorithm~\ref{alg:baseline-preselectors} lists the three baseline preselection snippets.  The Naive snippet returns the whole installed skill set.  The LLM preselection snippet first formats the full metadata list, then asks a selector model to name skills, and finally matches the returned names back to installed skills.  The RAG snippet embeds the instruction and skill metadata, ranks skills by embedding similarity, and returns the top-$k$ skills.

\begin{algorithm}[H]
\footnotesize
\caption{Baseline preselectors}
\label{alg:baseline-preselectors}
\begin{algorithmic}[1]
\REQUIRE instruction $I$, skills $S_s$, budget $k$, selector model $M_p$, embedding model $M_{emb}$
\ENSURE exposed skill set $P$
\STATE \textbf{Naive}$(I,S_s)$:
\STATE \quad \textbf{return} $S_s$
\STATE \textbf{LLMPreselect}$(I,S_s,k,M_p)$:
\STATE \quad $M \gets \textnormal{FormatMetadata}(S_s)$
\STATE \quad $N \gets M_p.\textnormal{SelectSkills}(I,M,k)$
\STATE \quad \textbf{return} $\textnormal{MatchSkills}(N,S_s)$
\STATE \textbf{RAGPreselect}$(I,S_s,k,M_{emb})$:
\STATE \quad $v_I \gets M_{emb}(I)$
\STATE \quad $E \gets \{(u, M_{emb}(m(u))) : u\in S_s\}$
\STATE \quad $R \gets \textnormal{RankBySim}(v_I,E)$
\STATE \quad \textbf{return} $\textnormal{TopK}(R,k)$
\end{algorithmic}
\end{algorithm}

\section{Prompt Templates}
\label{app:prompts}
This appendix gives the prompt templates. The execution agent receives skill metadata and must explicitly load a skill before using its longer instructions; preselection prompts instead return only skill names so that selected skills can be matched against the golden set.

\promptneedspace{10\baselineskip}
\subsection{Execution-agent initial prompt skeleton}
\begin{promptstrip}{Execution-agent initial prompt skeleton}
\begin{lstlisting}[style=promptstyle]
You are an AI assistant tasked with solving command-line tasks in a Linux environment. You will be given a task description and the output from previously executed commands. Your goal is to solve the task by providing batches of shell commands.

Format your response as JSON with the following structure:

{
  "analysis": "Analyze the current state based on the terminal output provided. What do you see? What has been accomplished? What still needs to be done?",
  "plan": "Describe your plan for the next steps. What commands will you run and why? Be specific about what you expect each command to accomplish.",
  "commands": [
    {"keystrokes": "ls -la\n", "duration": 0.1},
    {"keystrokes": "cd project\n", "duration": 0.1}
  ],
  "task_complete": true
}

Required fields:
- "analysis": Your analysis of the current situation
- "plan": Your plan for the next steps
- "commands": Array of command objects to execute

Optional fields:
- "task_complete": Boolean indicating if the task is complete (defaults to false if not present)

Command object structure:
- "keystrokes": String containing the exact keystrokes to send to the terminal (required)
- "duration": Number of seconds to wait for the command to complete (defaults to 1.0 if not present)

IMPORTANT: The text inside "keystrokes" will be used completely verbatim. Most bash commands should end with a newline (\n) to execute.

---
SKILL SYSTEM:
If skills are available (shown in "available_skills"), you MUST check if any skill matches the task BEFORE executing commands.

Skill context:
{skills_block}

To load a skill, use: cat <location> (where location is from the skill's <location> tag above)

LOADED SKILLS:
{active_skills}

To load a skill, respond with ONLY this JSON (no other fields):
{"load_skill": "skill-name"}

After a skill is loaded, proceed with the normal response format above.
Skills contain best-practice workflows - use them instead of writing custom solutions.
---

Task Description:
{instruction}

Current terminal state:
{terminal_state}
\end{lstlisting}
\end{promptstrip}

The execution prompt separates invocation from routing.  It can only load skills that appear in the exposed metadata block, so the preselection stage directly determines the action space visible to the agent.

\subsection{LLM preselection prompt}
\begin{promptstrip}{LLM preselection prompt}
\begin{lstlisting}[style=promptstyle]
You are a skill-routing assistant.
Given a task description and available skills,
select the skills most likely to be useful.

Return your answer as a JSON array of skill names. Return at most {max_skills}
skills. If no skill is relevant, return an empty array: []

IMPORTANT: Return ONLY the JSON array.
No explanation. No markdown code fences.

## Available Skills
{skills_list}

## Task Description
{task_instruction}

## Your Selection (JSON array of skill names)
\end{lstlisting}
\end{promptstrip}

This prompt uses a strict JSON-only output format, so selected skills can be matched unambiguously and explanations are never mistaken for skill names.

\promptneedspace{8\baselineskip}
\subsection{Hybrid preselection query-generation prompt}
\begin{promptstrip}{Hybrid preselection query-generation prompt}
\begin{lstlisting}[style=promptstyle]
You are a retrieval query planner for a skill-selection system.
Given a task description, write {num_queries}
short search queries covering distinct
capabilities likely needed for the task.

Guidelines:
- Each query should be concise, concrete, and skill-oriented.
- Prefer capability/library/workflow phrases.
- Do not restate the whole task.
- Return ONLY a JSON array of strings.
- No explanation. No markdown.

## Task Description
{task_instruction}

## Search Queries (JSON array)
\end{lstlisting}
\end{promptstrip}

The query-generation step is useful for multi-skill tasks because the original task instruction may emphasize the final artifact rather than every intermediate capability.

\subsection{Hybrid preselection LLM final selection prompt}
\begin{promptstrip}{RAG-to-LLM final selection prompt}
\begin{lstlisting}[style=promptstyle]
You are a skill-routing assistant.
Given a task description and available skills,
select the skills most likely to be useful.

Return your answer as a JSON array of skill names. Return at most {max_skills}
skills. If no skill is relevant, return an empty array: []

IMPORTANT: Return ONLY the JSON array.
No explanation. No markdown code fences.

## Available Skills
{skills_list}

## Task Description
{task_instruction}

## Your Selection (JSON array of skill names)
\end{lstlisting}
\end{promptstrip}

This final-selection prompt is the same as the LLM preselection prompt.

\subsection{Intervention prompts}
\label{app:intervention-prompts}
The field-level variants use the same prompt and replace only the designated distractor fields: name, description, location, or name plus location.  Unmasked fields and all golden-skill metadata remain readable.  The number of placeholders is proportional to the original field's character length, approximating rather than exactly matching its token length.  The field-level comparison is a targeted diagnostic rather than a test of statistical equivalence.

The masked intervention uses the same execution prompt as above, but the skill block keeps golden-skill metadata readable and replaces every irrelevant skill's name, description, and location with \texttt{<mask>} placeholders of approximately the original length.  A representative masked prompt is:
\begin{promptstrip}{Masked execution prompt}
\begin{lstlisting}[style=promptstyle]
You are an AI assistant tasked with solving command-line tasks in a Linux environment. You will be given a task description and the output from previously executed commands. Your goal is to solve the task by providing batches of shell commands.

Format your response as JSON with the following structure:
{ ... normal JSON response schema ... }

---
SKILL SYSTEM:
If skills are available (shown in "available_skills"), you MUST check if any skill matches the task BEFORE executing commands.

Skill context:
available_skills:
[
  {
    "name": "pid-controller",
    "description": "Implement PID control loops for control systems.",
    "location": "/root/.terminus/skills/pid-controller/SKILL.md"
  },
  {
    "name": "<mask> <mask>",
    "description": "<mask> <mask> <mask> <mask> <mask> <mask>",
    "location": "<mask> <mask> <mask> <mask> <mask>"
  },
  {
    "name": "<mask>",
    "description": "<mask> <mask> <mask> <mask> <mask> <mask> <mask>",
    "location": "<mask> <mask> <mask> <mask>"
  },
  ...
]

To load a skill, use: cat <location> (where location is from the skill's <location> tag above)

LOADED SKILLS:
{active_skills}

To load a skill, respond with ONLY this JSON (no other fields):
{"load_skill": "skill-name"}

After a skill is loaded, proceed with the normal response format above.
Skills contain best-practice workflows - use them instead of writing custom solutions.
---

Task Description:
{instruction}

Current terminal state:
{terminal_state}
\end{lstlisting}
\end{promptstrip}

The plan-first intervention keeps the original skill metadata and inserts one additional instruction immediately before the task description:
\promptneedspace{8\baselineskip}
\begin{promptstrip}{Plan-first execution prompt}
\begin{lstlisting}[style=promptstyle]
You are an AI assistant tasked with solving command-line tasks in a Linux environment. You will be given a task description and the output from previously executed commands. Your goal is to solve the task by providing batches of shell commands.

Format your response as JSON with the following structure:
{ ... normal JSON response schema ... }

---
SKILL SYSTEM:
If skills are available (shown in "available_skills"), you MUST check if any skill matches the task BEFORE executing commands.

Skill context:
{skills_block}

To load a skill, use: cat <location> (where location is from the skill's <location> tag above)

LOADED SKILLS:
{active_skills}

To load a skill, respond with ONLY this JSON (no other fields):
{"load_skill": "skill-name"}

After a skill is loaded, proceed with the normal response format above.
Skills contain best-practice workflows - use them instead of writing custom solutions.
---
\end{lstlisting}
\par\smallskip
\noindent\begin{minipage}{\linewidth}
\scriptsize\bfseries\linespread{1.15}\selectfont
IMPORTANT: Before executing any commands, you MUST first explicitly think about which of the available skills (if any) are relevant to this task. State your reasoning about skill relevance, then load the skills you need. Only after that should you begin solving the task.
\end{minipage}
\par\smallskip
\begin{lstlisting}[style=promptstyle]
Task Description:
{instruction}

Current terminal state:
{terminal_state}

\end{lstlisting}
\end{promptstrip}

\section{Use of AI Assistants}
\label{app:ai-assistants}
AI assistants were used during the preparation of this paper for writing support and for assistance with parts of the experimental code and analysis scripts.  The authors reviewed and edited the resulting text and code.

\section{Per-task Scaling Heatmaps}
\label{app:per-task-scaling-heatmaps}
\begin{figure}[!tp]
\centering
\includegraphics[width=\columnwidth,height=0.78\textheight,keepaspectratio]{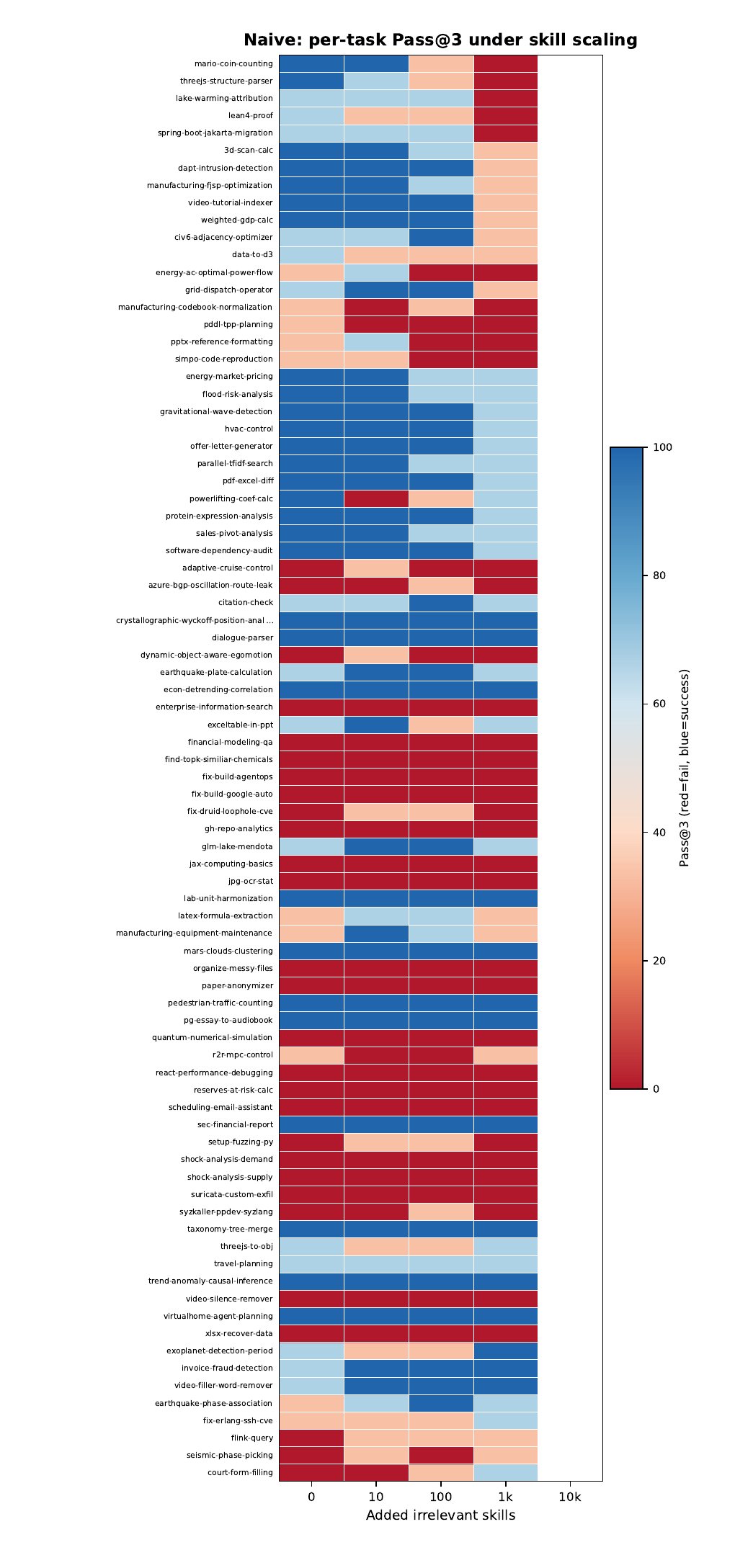}
\caption{Per-task Pass@3 for Naive exposure as the number of added irrelevant skills grows.  Red means all trials fail and blue means all trials pass.  Rows are tasks, sorted by the naive drop from no irrelevant skills to 1,000 irrelevant skills.}
\label{fig:task-pass3-heatmap-naive}
\end{figure}

\begin{figure}[!tp]
\centering
\includegraphics[width=\columnwidth,height=0.78\textheight,keepaspectratio]{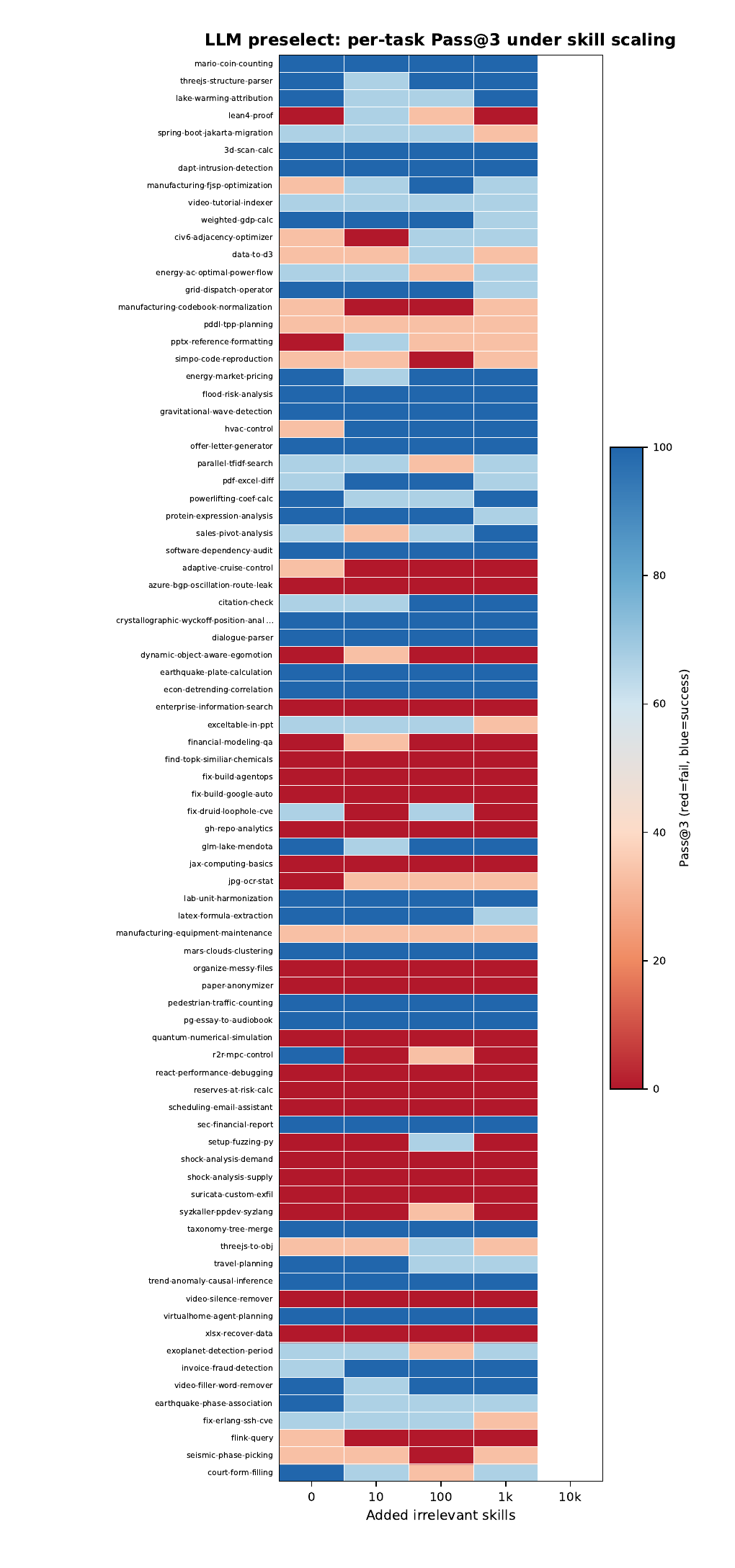}
\caption{Per-task Pass@3 for LLM preselection.  Compared with Naive exposure, many rows remain blue or light blue as the skill library grows, showing that explicit routing stabilizes task success when the full metadata list still fits in context.}
\label{fig:task-pass3-heatmap-llm}
\end{figure}

\begin{figure}[!tp]
\centering
\includegraphics[width=\columnwidth,height=0.78\textheight,keepaspectratio]{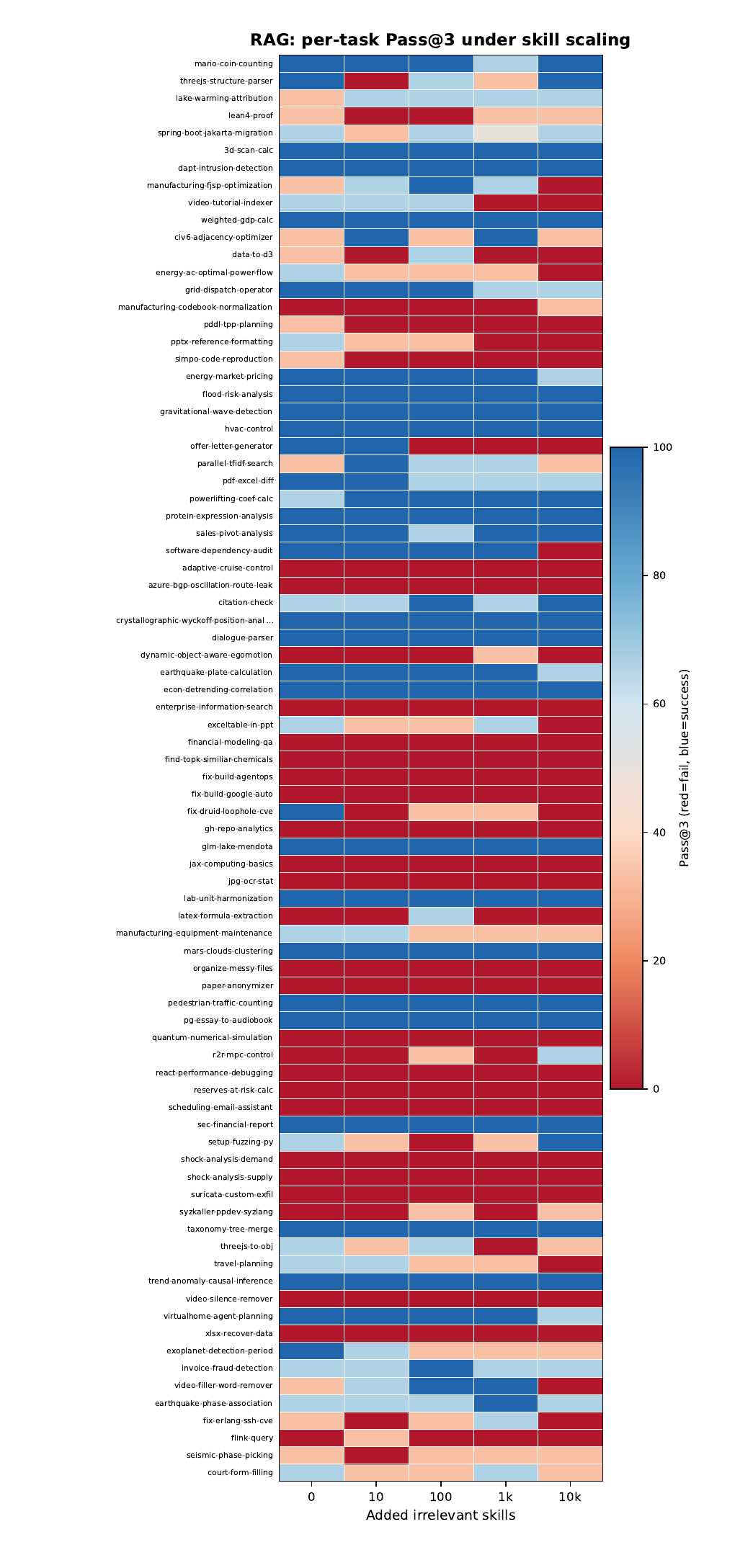}
\caption{Per-task Pass@3 for embedding RAG.  RAG is feasible at the 10k scale, but the mixed colors show that semantic retrieval alone misses some task-specific skill combinations.}
\label{fig:task-pass3-heatmap-rag}
\end{figure}

\begin{figure}[!tp]
\centering
\includegraphics[width=\columnwidth,height=0.78\textheight,keepaspectratio]{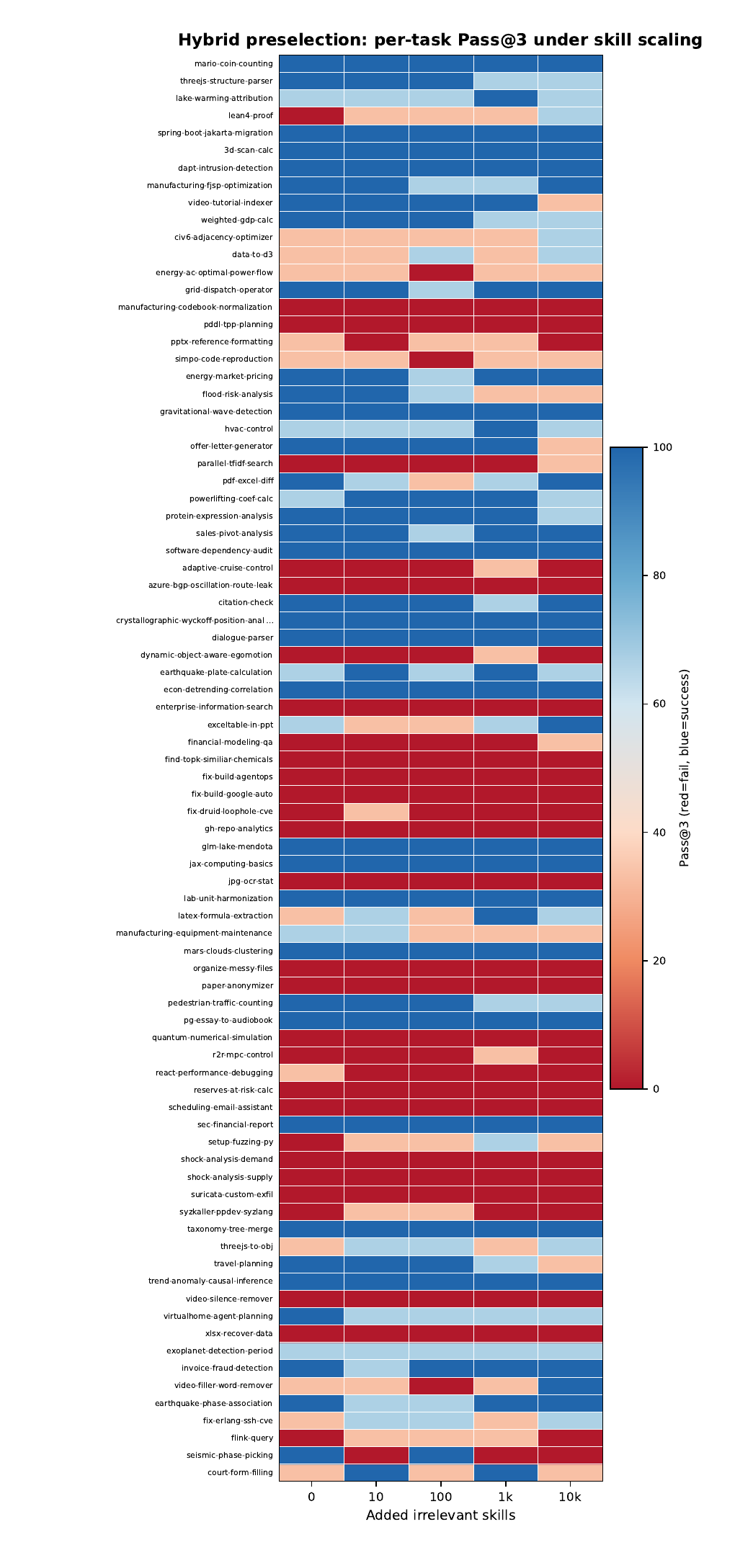}
\caption{Per-task Pass@3 for hybrid preselection.  The method keeps a compact candidate set while recovering more large-scale tasks than single-query RAG, especially when several capabilities must be retrieved together.}
\label{fig:task-pass3-heatmap-mq}
\end{figure}

The main text reports aggregate Pass@3.  This appendix adds a task-level view of that metric.  Figures~\ref{fig:task-pass3-heatmap-naive}--\ref{fig:task-pass3-heatmap-mq} show per-task Pass@3 as the number of added irrelevant skills increases.  Each row is one SkillsBench task, each column is a library-size setting, and the color encodes Pass@3 from red (all runs fail) to blue (all runs pass).

The heatmaps are sorted by the Naive drop from no irrelevant skills to 1,000 irrelevant skills, so collapses under implicit routing appear near the top.  This ordering makes the aggregate result more concrete: some tasks remain stable, but many tasks show abrupt failures once the library reaches hundreds or thousands of irrelevant skills.  The explicit-routing heatmaps then show whether those same rows become more stable when the execution agent sees a smaller selected skill set.

Figure~\ref{fig:task-pass3-heatmap-naive} shows the most direct form of routing collapse: rows that are blue or light blue at small scale often turn light red or red as irrelevant skills are added.  Figure~\ref{fig:task-pass3-heatmap-llm} shows the complementary result from the main text: when the full metadata list still fits, explicit LLM preselection keeps many of those rows more stable.  Figures~\ref{fig:task-pass3-heatmap-rag} and \ref{fig:task-pass3-heatmap-mq} focus on the scalable retrieval-based setting.  RAG remains feasible at 10,000 skills but produces more mixed task-level outcomes; hybrid preselection reduces some of this variance by constructing a more diverse candidate pool before final selection.

\end{document}